\documentclass[10pt,letterpaper,logo,twocolumn]{style}

\usepackage[numbers]{natbib}
\usepackage{graphicx}
\usepackage{booktabs}
\usepackage{amsmath,amsfonts,amssymb}
\usepackage{subcaption}
\usepackage{multirow}
\usepackage{colortbl}
\usepackage{listings}
\usepackage{xparse}
\usepackage{fontawesome5}
\usepackage{threeparttable}

\graphicspath{{./}{fig/}{figures/}{plot/}{pdf/}{table/}}
\usepackage{amsthm}
\usepackage{tcolorbox}
\tcbuselibrary{skins,breakable}
\tcbuselibrary{listingsutf8}
\usepackage{titletoc}
\usepackage{pifont}
\usepackage{mathtools}
\usepackage{bbm}
\usepackage{makecell}
\usepackage{adjustbox}
\usepackage{array}
\usepackage{tikz}
\usetikzlibrary{arrows.meta,positioning,calc,fit}
\usepackage[edges]{forest}

\definecolor{softsage}{RGB}{189,205,179}
\definecolor{softsky}{RGB}{173,198,222}
\definecolor{softpurple}{RGB}{198,188,214}
\definecolor{softgray}{RGB}{230,230,230}

\providecommand{\IEEEPARstart}[2]{#1#2}
\providecommand{\appendices}{\appendix}
\providecommand{\shortcite}[1]{\cite{#1}}

\newcommand{\githubartifact}[1]{\href{#1}{\faIcon{github}}}
\newcommand{\hfdataset}[1]{\href{#1}{\faIcon{database}}}
\newcommand{\webartifact}[1]{\href{#1}{\faIcon{link}}}

\newcounter{researchquestion}
\makeatletter
\@addtoreset{researchquestion}{subsection}
\makeatother
\renewcommand{\theresearchquestion}{\arabic{researchquestion}}

\newcommand{\researchquestion}[1]{%
  \refstepcounter{researchquestion}%
  \par\addvspace{\medskipamount}%
  \noindent
  \begin{tcolorbox}[
    enhanced,
    colback=softsky!10,
    colframe=black!55,
    borderline west={2.5pt}{0pt}{softsky!75!black},
    boxrule=0.45pt,
    arc=1.5pt,
    left=7pt,
    right=7pt,
    top=4pt,
    bottom=4pt,
    boxsep=0pt,
    before skip=0pt,
    after skip=0pt,
    nobeforeafter,
    width=\linewidth,
  ]
  {\theresearchquestion.\ #1}
  \end{tcolorbox}%
  \nopagebreak
  \vspace{0.2\baselineskip}}

\newtcolorbox{definitionbox}{
  colback=softgray!25,
  colframe=black!55,
  boxrule=0.45pt,
  arc=1.5pt,
  left=6pt,
  right=6pt,
  top=5pt,
  bottom=5pt,
  before skip=0.75\baselineskip,
  after skip=0.75\baselineskip
}

\title{A Survey on \textit{Agent-as-a-Judge}}
\runningtitle{A Survey on Agent-as-a-Judge}
\PublicDate{2026-11-12}

\makeatletter
\renewcommand{\AB@authnote}[1]{}
\makeatother
\author{%
  \textbf{Runyang~You}\textsuperscript{1}\equal,\ 
  \textbf{Hongru~Cai}\textsuperscript{1}\equal,\ 
  \textbf{Caiqi~Zhang}\textsuperscript{2},\ 
  \textbf{Qiancheng~Xu}\textsuperscript{1},\ 
  \textbf{Meng~Liu}\textsuperscript{3},\ 
  \textbf{Tiezheng~Yu}\textsuperscript{4},\ 
  \textbf{Yongqi~Li}\textsuperscript{1}\advisor, and
  \textbf{Wenjie~Li}\textsuperscript{1}%
  \\
  \Affilfont
  \textsuperscript{1}The Hong Kong Polytechnic University \quad
  \textsuperscript{2}University of Cambridge%
  \\
  \Affilfont
  \textsuperscript{3}Shandong University \quad
  \textsuperscript{4}Huawei~Technologies%
  \\
  \Affilfont\texttt{runyang.y@outlook.com, \{henry.hongrucai, liyongqi0\}@gmail.com}%
}

\begin{document}

\begin{abstract}
Artificial intelligence has progressed from task-specific systems toward general models spanning perception, generation, reasoning, and autonomous interaction, making reliable evaluation essential for comparison, deployment, and post-training feedback.
At scale, modern evaluation pipelines increasingly rely on \textit{LLM-as-a-Judge} to produce scores, preferences, and critiques; in its fixed-context setting, the judge relies on the supplied evaluand and evaluation materials.
Yet a reliable verdict may require evidence beyond the supplied evaluand and context---either because critical environmental states remain unrecorded in the static execution log, or because controversial factual claims require external verification.
Passive judges cannot obtain that missing evidence before deciding, especially when evaluating agent trajectories, executable artifacts, or domain-specific reports.
Recent work therefore develops active evaluators that seek missing evidence and update their grounds for judgment before deciding.
We bring these advances together through \textit{Agent-as-a-Judge}, focusing on how evaluators construct and control checking processes within the broader LLM-based judging landscape.
We define the scope through evaluation-directed checks and the use of their results, covering both structured workflows and observation-driven control.
We organize the design space by multi-agent collaboration, tool integration, memory, and planning, and show how these mechanisms supply the checks and evidence on which a verdict rests.
We consolidate published scores and reproduce missing entries for key agentic judges;
the comparison reveals uneven agentic gains and motivates dedicated agentic evaluation.
We conclude by characterizing application demands and open challenges in cost, reliability, safety, and governance that condition trusted deployment of agentic judges at scale.

\end{abstract}

\newcommand{\TitleLinks}{%
\centering
    \vspace{6pt}
    {\noindent\absfont\fontsize{11}{13}\selectfont
    \faGithub\ Project Page: \url{https://github.com/ModalityDance/Awesome-Agent-as-a-Judge}\par}%
}

\begin{teaserfigure}
  \includegraphics[width=\textwidth]{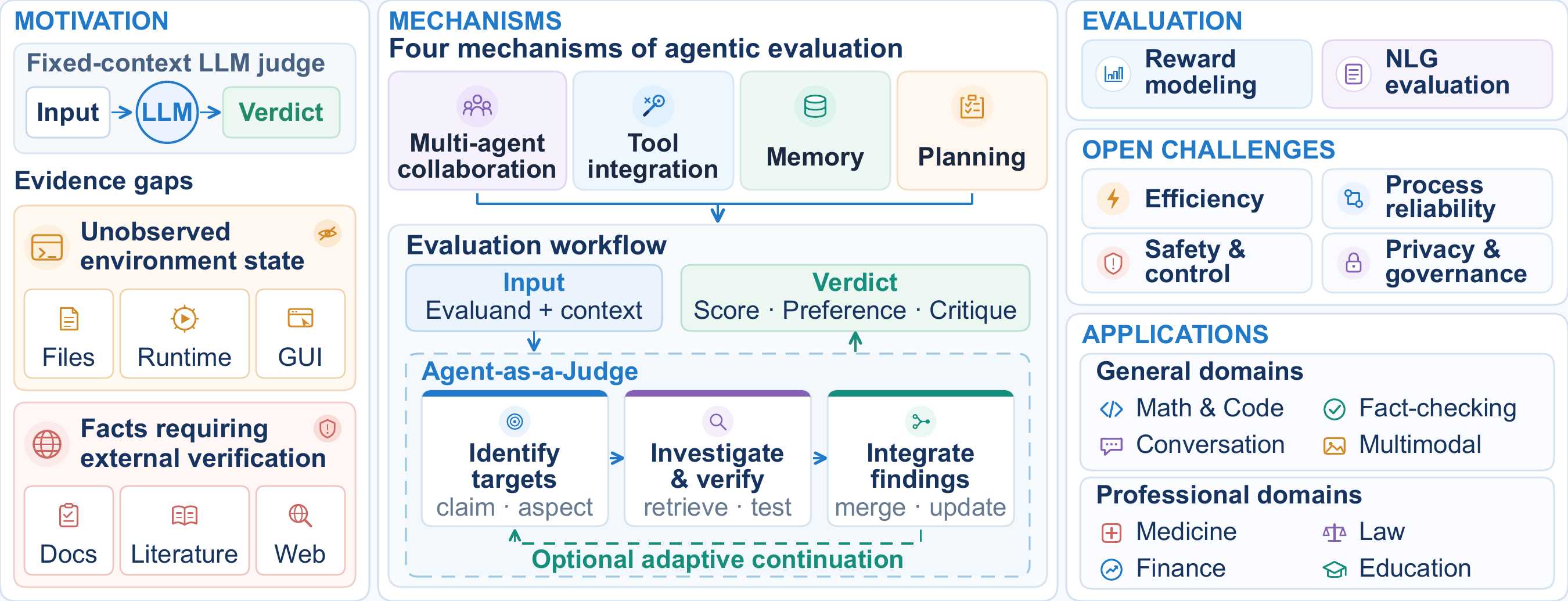}
  \caption{
  Conceptual landscape of \textit{Agent-as-a-Judge}.
  Unlike fixed-context LLM judges, an \textit{agentic judge} resolves missing evidence through evaluation-directed checking---identifying targets, verifying via tools, and integrating findings into grounded verdicts---driven by four mechanisms (multi-agent collaboration, tool integration, memory, planning) across evaluation paradigms, open challenges, and target domains.
  }
  \label{fig:teaser}
\end{teaserfigure}

\setlength{\TitleFigureSep}{4pt}
\setlength{\TitleBodySep}{2pt}
\maketitle

\section{Introduction}

\begin{figure}[t]
    \centering
    \begin{subfigure}[b]{0.49\linewidth}
        \includegraphics[width=\textwidth]{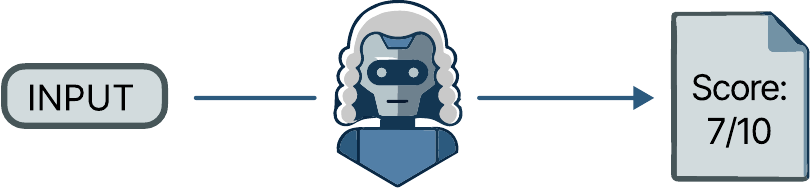}
        \caption{LLM-as-a-Judge}
        \label{subfig:llm-as-a-judge}
    \end{subfigure}
    \\
    \begin{subfigure}[b]{0.85\linewidth}
        \includegraphics[width=\textwidth]{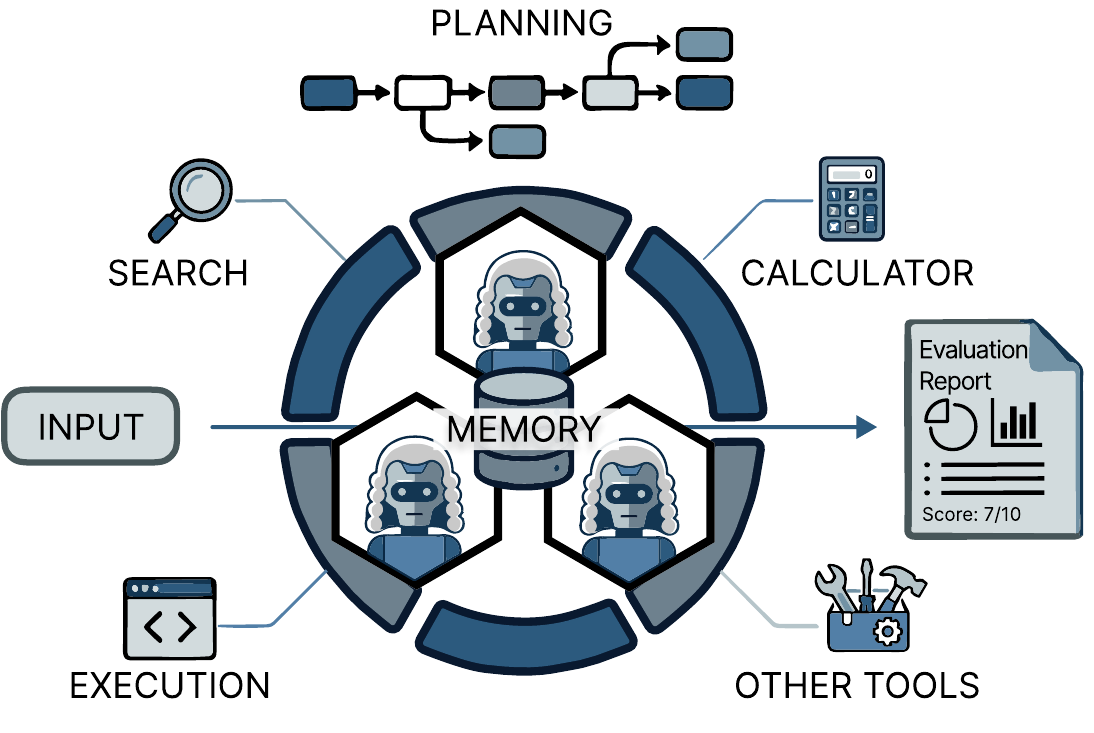}
        \caption{Agent-as-a-Judge}
        \label{subfig:agent-as-a-judge}
    \end{subfigure}
    \caption{
    Comparison between (a) LLM-as-a-Judge and (b) Agent-as-a-Judge.
    The former acts as a passive evaluator relying on the given context and model-internal knowledge, while the latter acts as an active evaluator that plans, gathers evidence, uses tools, and iteratively verifies its judgment.
    }
    \label{fig:paradigm-comparison}
\end{figure}

\IEEEPARstart{A}{rtificial} intelligence has progressed from task-specific systems toward general models spanning perception, generation, reasoning, and autonomous interaction, making reliable evaluation essential for comparison, deployment, and post-training feedback~\citep{christiano2017deep,stiennon2020learning,lai2025post-training-scaling}.
Human judgment has long served as the reference standard for these evaluations~\citep{christiano2017deep,stiennon2020learning,fabbri2021summeval,wang-etal-2023-automated},
yet manual annotation is costly and slow, and inter-annotator disagreement limits consistency and reliability~\citep{stiennon2020learning,ouyang2022training,fabbri2021summeval,li2025LLM-as-a-Judge-Survey}---limits that become especially acute when evaluation must run continuously during post-training or across large-scale public leaderboards~\citep{chiang2024chatbot,lai2025post-training-scaling}.
To improve scalability, the field adopted automatic metrics that score outputs against predefined rules or references~\citep{papineni2002bleu,lin2004rouge,rajpurkar2016squad}.
These approaches are efficient and reproducible, but they presuppose tasks with formalizable targets and struggle with open-ended generation, fine-grained quality, and judgments that depend on broader context~\citep{fabbri2021summeval,wang-etal-2023-automated,li2025LLM-as-a-Judge-Survey}.

Rapid advances in large language models (LLMs) have opened a third path~\citep{li2025LLM-as-a-Judge-Survey}.
Modern LLMs combine strong semantic understanding with instruction-following capabilities, allowing them to follow evaluation prompts and provide human-like quality judgments~\citep{ouyang2022training,Zheng2023Judging,liu2023G-Eval,kim2023Prometheus,li2025LLM-as-a-Judge-Survey}.
In the fixed-context setting shown in Figure~\ref{subfig:llm-as-a-judge}, a judge reads a supplied evaluand---the object under evaluation, such as a response, summary, or trajectory---together with its context and returns a score, preference, or critique~\citep{Zheng2023Judging,liu2023G-Eval}.
A judge LLM can evaluate large volumes of outputs via repeated inference at machine speed and cost, while maintaining near-human judgment quality on many open-ended tasks~\citep{Zheng2023Judging,liu2023G-Eval,zhu2025judgelm}.
Collectively, these properties make LLM judges a scalable proxy for human preference~\citep{li2025LLM-as-a-Judge-Survey}, whose judgments now underpin model comparison and scoring, reward signals for reinforcement learning~\citep{lee2024RLAIF-vs-RLHF}, and large-scale synthetic data curation~\citep{long2024llmsdatageneration} --- placing judge fidelity at the center of modern evaluation and optimization pipelines~\citep{lai2025post-training-scaling}.

However, this passive design proves insufficient once a reliable verdict depends on evidence beyond the materials provided with the case~\citep{zhuge2025AgentasaJudge,peng2025AgenticRM,zhang2025EvaluationAgent}.
What is missing may be an unobserved environment status, such as code execution outcomes, workspace state, and live web pages; or facts that fall outside the judge's knowledge and must be recovered from external sources~\citep{zhuge2025AgentasaJudge,shi2026AJBench,peng2025AgenticRM,wang2026ToA}.
Obtaining them typically requires retrieval, re-execution, or environment interaction before a reliable score, preference, or critique can be issued~\citep{peng2025AgenticRM,ahmad-etal-2025-urdufactcheck,zhang2025EvaluationAgent}.
The gap is especially visible when the evaluand itself is an agent or a trajectory from an agent, where success depends on external state and procedure as well as the textual response to user~\citep{zhuge2025AgentasaJudge,zhang2025EvaluationAgent,jeong-etal-2025-agent,ghosh2024SAGEval,xi2025LLM-agent-survey}.
Under passive \textit{LLM-as-a-Judge}, however, the judge scores only from that fixed packet and cannot gather the missing but decisive evidence before issuing a verdict~\citep{zhuge2025AgentasaJudge,zhang2025EvaluationAgent,lin-etal-2025-fact,peng2025AgenticRM}.
Reliable judgement accordingly requires an active judge that can gather and update evidence beyond what was initially provided~\citep{peng2025AgenticRM,zhang2025EvaluationAgent}.

The emerging literature answers with \textit{Agent-as-a-Judge} (Figure~\ref{subfig:agent-as-a-judge}):
an evaluator that maintains an evaluation state, carries out checks, and incorporates their results before issuing a verdict~\citep{zhuge2025AgentasaJudge,peng2025AgenticRM,zhang2025EvaluationAgent}.
Mechanically, 
tool integration lets judges obtain external evidence~\citep{peng2025AgenticRM,xu2025TIRJudge,han2025VerifiAgent}; planning guides how claims are inspected and verified as that evidence accumulates~\citep{zhang2025EvaluationAgent,wadhwa2025EvalAgents}; memory retains findings, case artifacts, and curated evaluation experience across steps~\citep{zhao2025RLPA,ding2025ARMThinker,lin-etal-2025-fact,yu2025TableCritic}.
Multi-agent collaboration further mitigates single-evaluator bias by distributing criteria, surfacing objections, and aggregating independent assessments~\citep{chan2024ChatEval,chen2025MultiAgentasJudge,jiang2025agentsbench}.
Together, these mechanisms extend what a judge can inspect, ground verdicts in traceable evidence, and sustain evaluation across steps---
positioning agentic judges as an increasingly central approach to reliable evaluation.

Existing surveys provide complementary foundations: Li et al.~\citep{li2025LLM-as-a-Judge-Survey} organize LLM-based judgment around what to judge, how to judge, and how to benchmark, while Yehudai et al.~\citep{yehudai2026agentEvaluationSurvey} map agent capabilities, benchmarks, and evaluation frameworks.
Our organizing object is the evaluator's own checking process: how it obtains findings, connects intermediate results, and controls verification before forming a verdict.
We bring work across training pipelines, verification tasks, evaluator architectures, and domains into a survey of \textit{Agent-as-a-Judge} centered on these mechanisms and their meta-evaluation.
Building on this view, our contributions are as follows:
\begin{itemize}
    \item An operational definition and scope for \textit{Agent-as-a-Judge}:
    we characterize evaluation-directed checks and the use of their results, distinguishing these processes from fixed-context scoring and replicated judgments,
    and describe control from prescribed workflows to observation-driven adaptation (Section~\ref{sec:background}).
    \item A mechanism taxonomy over multi-agent collaboration, tool integration, memory, and planning:
    we organize the literature under this taxonomy and show how these mechanisms supply the checks and evidence on which a verdict rests (Section~\ref{sec:methods}).
    \item A comparative meta evaluation of agentic judges:
    we consolidate published scores and reproduce missing entries for five key agentic judges across three unified benchmarks;
    the comparison shows uneven agentic gains under preference-oriented benchmarks and motivates  dedicated evaluation of evidence-seeking reliability (Section~\ref{sec:evaluation}).
    \item An account of application demands together with open challenges in cost, reliability, safety, and governance that condition trusted deployment of agentic judges at scale (Sections~\ref{sec:apps}--\ref{sec:discussion}).
\end{itemize}

\begin{figure*}[t]

    \pgfkeys{/forest/col1/.style={text width=1.2cm}}
    \pgfkeys{/forest/col2/.style={text width=2.5cm}}
    \pgfkeys{/forest/col3/.style={text width=2.6cm}}
    \pgfkeys{/forest/colContent/.style={text width=6.1cm}}
    \pgfkeys{/forest/colWideContent/.style={text width=9cm}}
    \pgfkeys{/forest/axisApps/.style={fill=softsage!60, draw=softsage!85!black}}
    \pgfkeys{/forest/axisFunc/.style={fill=softsky!60, draw=softsky!85!black}}
    \pgfkeys{/forest/axisMethod/.style={fill=softpurple!60, draw=softpurple!85!black}}
    \pgfkeys{/forest/subApps/.style={fill=softsage!38, draw=softsage!70!black}}
    \pgfkeys{/forest/subFunc/.style={fill=softsky!38, draw=softsky!70!black}}
    \pgfkeys{/forest/subMethod/.style={fill=softpurple!38, draw=softpurple!70!black}}
    \pgfkeys{/forest/leafApps/.style={fill=softsage!24, draw=softsage!55!black}}
    \pgfkeys{/forest/contentApps/.style={fill=softsage!14, draw=softgray}}
    \pgfkeys{/forest/contentFunc/.style={fill=softsky!14, draw=softgray}}
    \pgfkeys{/forest/contentMethod/.style={fill=softpurple!14, draw=softgray}}
    \resizebox{\textwidth}{!}{
    \begin{forest}
    for tree={   
        font=\fontsize{6}{5}\selectfont,
        draw=softgray, semithick, rounded corners,
        minimum height = 1.ex,
        minimum width = 2em,
        anchor = west,
        grow = east,
        forked edge,        
        s sep = 0.5mm,    
        l sep = 2mm,    
        fork sep = 1mm,            
    }
    [Agent-as-a-Judge, anchor=center, rotate=90,
        fill=white, draw=softgray!80!black, line width=0.7pt,
        font=\fontsize{7}{6}\selectfont
        [Application (\S \ref{sec:apps}), fit=band, col1, axisApps
            [Professional Domains \\(\S \ref{sec:apps:professional}), col2, subApps
                [Education, col3, leafApps, l sep = 1mm
                    [{E.g.,
                    AutoSCORE~\shortcite{wang2025autoscore},
                    GradeOpt~\shortcite{chu2024llm},
                    Grade-Like-a-Human~\shortcite{xie2024grade}
                    }, colContent, contentApps]
                ]
                [Finance, col3, leafApps, l sep = 1mm
                    [{E.g.,
                    FinResearchBench~\shortcite{sun2025finresearchbench},
                    FinDeepResearch~\shortcite{zhu2025findeepresearch},
                    SAEA~\shortcite{chen2025standardbenchmarksfail},
                    M-SAEA~\shortcite{chen2025tasks}
                    }, colContent, contentApps]
                ]
                [Law, col3, leafApps, l sep = 1mm
                    [{E.g., AgentsCourt~\shortcite{he2024agentscourt}, 
                    SAMVAD~\shortcite{devadiga2025samvad},
                    AgentsBench~\shortcite{jiang2025agentsbench}
                    }, colContent, contentApps]
                ]
                [Medicine, col3, leafApps, l sep = 1mm
                    [{E.g.,
                    GEMA-Score~\shortcite{zhenxuan2025GEMA-Score},
                    ChatCoach~\shortcite{huang2024benchmarking},
                    AI Hospital~\shortcite{fan2025ai}
                    }, colContent, contentApps]
                ]
            ]
            [General Domains \\(\S \ref{sec:apps:general}), col2, subApps
                [Multimodal and Vision, col3, leafApps, l sep = 1mm
                    [{E.g., CIGEval~\shortcite{wang-etal-2025-unified}, 
                    LRQ-Fact~\shortcite{beigi2024lrqfact}
                    }, colContent, contentApps]
                ]
                [Conversation and Interaction, col3, leafApps, l sep = 1mm
                    [{E.g., IntellAgent~\shortcite{levi2025intellagent}, 
                    ESC-Judge~\shortcite{madani-srihari-2025-esc}, 
                    Sentient Agent~\shortcite{zhang2025sentient}, 
                    PSYCHE~\shortcite{lee2025psyche}
                    }, colContent, contentApps]
                ]
                [Fact-Checking, col3, leafApps, l sep = 1mm
                    [{E.g., FACT-AUDIT~\shortcite{lin-etal-2025-fact}, 
                    UrduFactCheck~\shortcite{ahmad-etal-2025-urdufactcheck}, 
                    NarrativeFactScore~\shortcite{jeong-etal-2025-agent}, 
                    }, colContent, contentApps]
                ]
                [Math and Code, col3, leafApps, l sep = 1mm
                    [{E.g.,
                    CompassVerifier~\shortcite{liu-etal-2025-compassverifier}, 
                    xVerify~\shortcite{chen2025xverify}, 
                    Popper~\shortcite{huang2025popper}
                    }, colContent, contentApps]
                ]
            ]
        ]
        [Evaluation \\(\S \ref{sec:evaluation}), fit=band, col1, axisFunc
            [NLG Evaluation \\(\S \ref{sec:evaluation-nlg}), col2, subFunc, l sep = 2mm
                [{E.g.,
                ChatEval~\shortcite{chan2024ChatEval},
                MAJ-EVAL~\shortcite{chen2025MultiAgentasJudge},
                CollabEval~\shortcite{gupta2026CollabEval},
                MATEval~\shortcite{Li2024MATEval},
                SAGEval~\shortcite{ghosh2024SAGEval},
                CourtEval~\shortcite{kumar2025courteval}
                }, colWideContent, contentFunc]
            ]
            [Reward Modeling \\(\S \ref{sec:evaluation-rm}), col2, subFunc, l sep = 2mm
                [{E.g.,
                TIR-Judge~\shortcite{xu2025TIRJudge},
                Agentic Reward Model~\shortcite{peng2025AgenticRM},
                Multi-Agent Verification~\shortcite{lifshitz2025multi},
                VerifiAgent~\shortcite{han2025VerifiAgent},
                Themis~\shortcite{li2024toolaugmented}
                }, colWideContent, contentFunc]
            ]
        ]
        [Methodology \\(\S \ref{sec:methods}), fit=band, col1, axisMethod
            [Planning \\(\S \ref{sec:plan}), col2, subMethod, l sep = 2mm
            [{E.g.,
            Evaluation Agent~\shortcite{zhang2025EvaluationAgent},
            EvalAgents~\shortcite{wadhwa2025EvalAgents},
            AGENT-X~\shortcite{li2025AGENT-X},
            ARJudge~\shortcite{xu2025ARJudge},
            OnlineRubrics~\shortcite{rezaei2025OnlineRubrics}
            }, 
            colWideContent,
            contentMethod]
            ]
            [Memory \\(\S \ref{sec:memory}), col2, subMethod, l sep = 2mm
            [{E.g.,
            ARM-Thinker~\shortcite{ding2025ARMThinker}, 
            RLPA~\shortcite{zhao2025RLPA},
            SynthesizeMe~\shortcite{ryan2025SynthesizeMe},
            PersRM-R1~\shortcite{li2025PersRMR1},
            FSPO~\shortcite{singh2025FSPO}
            }, colWideContent, contentMethod]
            ]
            [Tool Integration \\(\S \ref{sec:tool}), col2, subMethod, l sep = 2mm
                [{E.g., Agent-as-a-Judge~\shortcite{zhuge2025AgentasaJudge},
                CodeVisionary~\shortcite{wang2025CodeVisionary},
                HERMES~\shortcite{ospanov2025HERMES}
                }, colWideContent, contentMethod]
            ]
            [Multi-Agent Collaboration \\(\S \ref{sec:multiagent}), col2, subMethod, l sep = 2mm
                [{E.g.,
                M-MAD~\shortcite{feng2025MMAD},
                HiMATE~\shortcite{zhang2025HiMATE},
                CAFES~\shortcite{su2025CAFES}
                }, 
                colWideContent,
                contentMethod]
            ]
        ]
    ]
    \end{forest}
    }
    \caption{A taxonomy of Agent-as-a-Judge organizing Methodology (\S \ref{sec:methods}), Evaluation (\S \ref{sec:evaluation}), and Applications (\S \ref{sec:apps}).}
    \label{fig:taxonomy}
    \end{figure*}
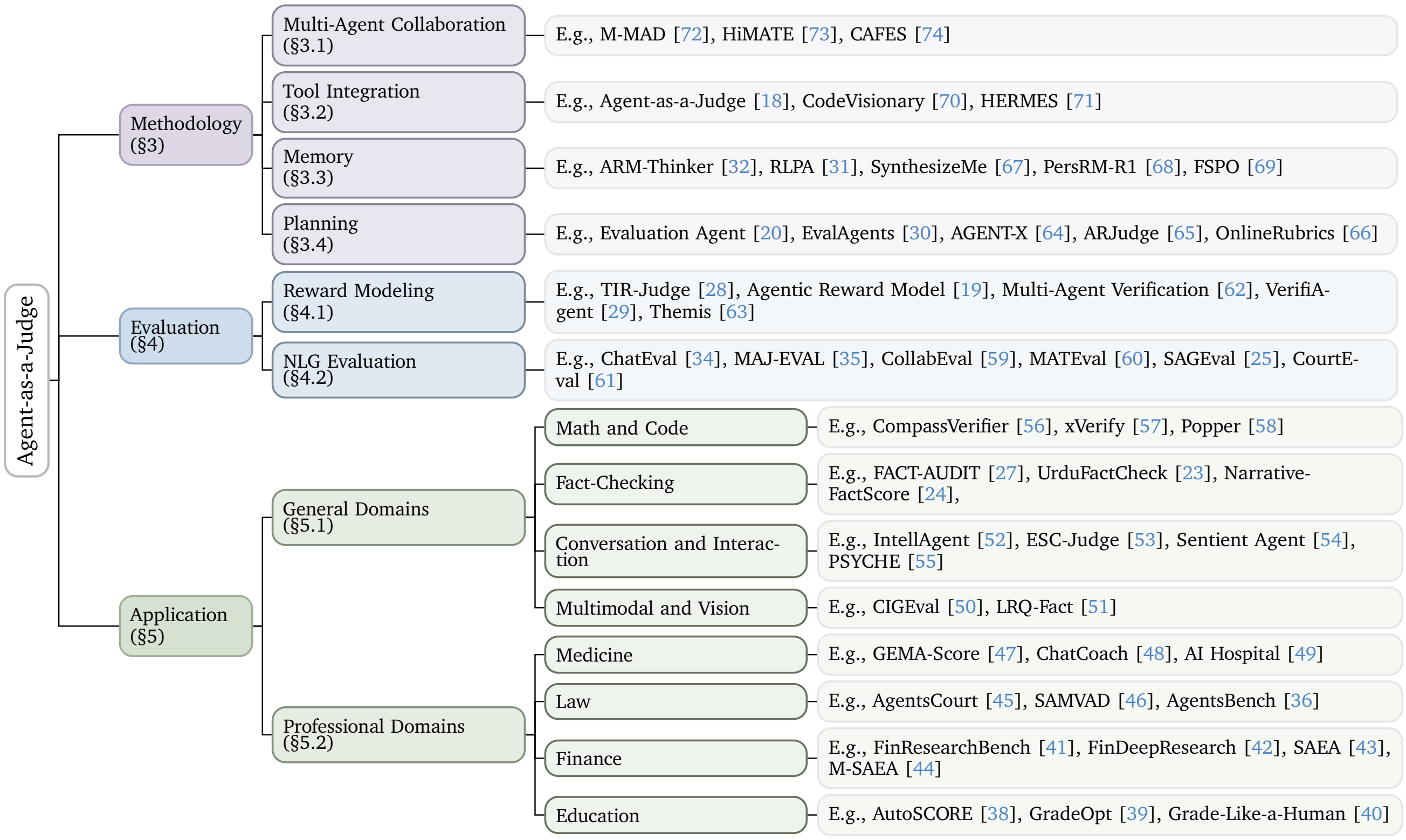



\section{Background}
\label{sec:background}
Within the broader \textit{LLM-as-a-Judge} landscape, we focus on evaluators that construct judgments through explicit checking processes.
We first formalize fixed-context judging as a reference setting, then examine the need for additional evidence and introduce the operational scope of \textit{Agent-as-a-Judge}.

\subsection{\textit{LLM-as-a-Judge}}

\textit{LLM-as-a-Judge} emerged as a pragmatic response to a long-standing dilemma in evaluation.
From the outset, human judges have assessed quality through pairwise preferences, Likert-scale ratings, and expert rubrics,
serving as the reference standard for semantic nuance and contextual appropriateness~\citep{christiano2017deep,stiennon2020learning,chiang2024chatbot,Zheng2023Judging,fabbri2021summeval,wang-etal-2023-automated}.
Yet manual annotation is costly and slow,
and inter-annotator disagreement limits reproducibility when evaluation must run continuously during post-training or across large public leaderboards~\citep{stiennon2020learning,ouyang2022training,chiang2024chatbot,li2025LLM-as-a-Judge-Survey}.
Automatic metrics were introduced to improve scalability: they score outputs against predefined rules or references,
such as BLEU~\citep{papineni2002bleu} and ROUGE~\citep{lin2004rouge} for translation and summarization,
and exact-match or F1 scores in question answering~\citep{rajpurkar2016squad}.
These approaches are efficient and reproducible,
but they presuppose tasks with formalizable targets and struggle with open-ended generation, fine-grained quality,
and judgments that depend on broader context~\citep{fabbri2021summeval,wang-etal-2023-automated,li2025LLM-as-a-Judge-Survey}.
Human evaluation and automatic metrics thus trade semantic judgment quality against deployment scale.

Large language models opened a third path~\citep{li2025LLM-as-a-Judge-Survey}.
Because they follow natural-language instructions and rubrics, they can return scores, preferences, or critiques on open-ended outputs rather than matching fixed surface patterns~\citep{ouyang2022training,Zheng2023Judging,liu2023G-Eval,kim2023Prometheus},
while repeated inference applies the same evaluation prompt to arbitrarily many evaluands at machine speed and cost~\citep{Zheng2023Judging,zhu2025judgelm}.
\textit{LLM-as-a-Judge} therefore targets the semantic flexibility missing from automatic metrics without reintroducing the annotation burden of human evaluation~\citep{Zheng2023Judging,liu2023G-Eval,kim2023Prometheus}.

Early work on passive \textit{LLM-as-a-Judge} mainly followed two lines~\citep{li2025LLM-as-a-Judge-Survey}.
One concerns judgment format: pairwise preference compares two candidates---automated in MT-Bench and Chatbot Arena~\citep{Zheng2023Judging,chiang2024chatbot} and reused as alignment rewards~\citep{ouyang2022training,lee2024RLAIF-vs-RLHF}---while pointwise scoring assigns an absolute grade to each response~\citep{Zheng2023Judging,zhu2025judgelm}.
The other enriches the supplied context $\mathcal{C}$, typically with explicit rubrics that structure criteria when gold references are absent or insufficient~\citep{chen2026rubrics}, as in G-Eval~\citep{liu2023G-Eval} and Prometheus~\citep{kim2023Prometheus}.
Across both lines, the judge still receives a fixed evaluand and context and returns a verdict without actively expanding the available evidence (Figure~\ref{fig:paradigm-comparison}).

\begin{definitionbox}
\textbf{Definition (\textit{LLM-as-a-Judge}).}
In the fixed-context setting considered here, an \textit{LLM-as-a-Judge} maps an evaluand and a supplied evaluation context to a verdict:
\[
\hat{y} = J_{\mathrm{LLM}}(x, \mathcal{C}),
\]
where $x$ is the evaluand, $\mathcal{C}$ is the supplied evaluation context, and $\hat{y}$ is the verdict.
\end{definitionbox}

This passive formulation treats the supplied evaluand $x$ and context $\mathcal{C}$---rubrics, references, demonstrations, partial traces---as the complete evidence available at verdict time,
as in Figure~\ref{subfig:llm-as-a-judge}.
Once passive LLM judges proved deployable at scale with sufficient alignment to human judgments on many open-ended tasks~\citep{Zheng2023Judging,liu2023G-Eval,li2025LLM-as-a-Judge-Survey},
their verdicts were adopted for pairwise model comparison, reinforcement-learning reward modeling~\citep{lee2024RLAIF-vs-RLHF}, synthetic-data filtering~\citep{long2024llmsdatageneration}, and broader post-training pipelines~\citep{lai2025post-training-scaling}.
This passive contract remains well matched when the evaluand is an open-ended model response,
but it is strained once the evaluand evolves further---to agent trajectories and tool-mediated workflows, or to domain reports whose claims must be checked against external state or references~\citep{xi2025LLM-agent-survey,zhuge2025AgentasaJudge,peng2025AgenticRM,ahmad-etal-2025-urdufactcheck}.

\subsection{From \textit{LLM-as-a-Judge} to \textit{Agent-as-a-Judge}}

Under such evaluands, a reliable verdict may depend on evidence that the closed pair $(x,\mathcal{C})$ does not contain.
The failure takes two related forms: decisive evidence about external state that is absent from the supplied text, and an epistemic gap when needed facts fall outside the judge's parametric knowledge.

First, decisive evidence may be absent from the text of $x$ itself---for example when $x$ is an agent trajectory, a code patch that must be verified by execution, or a computer-use trace whose effects on files, runtime state, or the GUI are not fully captured in the supplied text~\citep{xi2025LLM-agent-survey,zhuge2025AgentasaJudge,shi2026AJBench}.
Workspace files, database records, GUI state, execution results, and live web content sit outside what a passive judge can read from the supplied context~\citep{zhuge2025AgentasaJudge,shi2026AJBench,zhang2025EvaluationAgent}.
An evaluand that looks complete in text may still be inconsistent with that external state,
so verification requires obtaining observations---re-reading files, querying state, re-executing critical steps---before the verdict~\citep{zhuge2025AgentasaJudge,shi2026AJBench,peng2025AgenticRM}.
A passive judge confined to $(x,\mathcal{C})$ cannot produce those external observations~\citep{zhuge2025AgentasaJudge,peng2025AgenticRM}.

Second, critical facts may be absent from the context and the judge's parametric knowledge.
Whether a claimed dosage matches current prescribing information,
whether a clinical recommendation accords with updated guidelines,
or whether a scientific statement is supported by the cited primary literature
depends on external references that the supplied materials may not contain~\citep{peng2025AgenticRM,shi2026AJBench,ahmad-etal-2025-urdufactcheck}.
When those facts cannot be recovered reliably through internal reasoning alone,
external evidence-seeking is necessary for the judge~\citep{wang2026ToA};
otherwise a closed $(x,\mathcal{C})$ forces the verdict onto the judge's priors
and may penalize novel yet valid approaches that the judge simply does not know.

Both failures are instances of the same closed-world limit:
$(x,\mathcal{C})$ lacks evidence the verdict needs,
whether that evidence is an environment observation or external fact beyond the judge's knowledge.
Reliable judgment therefore requires an evaluator that can seek the missing evidence before deciding---
querying environment state when success is not visible in the supplied text,
and recovering documentation or other external facts when the needed knowledge cannot be settled internally~\citep{peng2025AgenticRM,zhang2025EvaluationAgent,shi2026AJBench,wang2026ToA}.
This motivates \textit{Agent-as-a-Judge}: evaluation processes that construct their grounds for judgment through evidence acquisition and explicit checking.

\subsection{\textit{Agent-as-a-Judge}}
Figure~\ref{subfig:agent-as-a-judge} illustrates the contrast with passive \textit{LLM-as-a-Judge}.
An agentic judge organizes judgment as an explicit evaluation process: it carries out checks, accumulates findings, and grounds its verdict in their results.
Tool integration, planning, memory, and multi-agent collaboration provide the principal mechanisms for building this process.

\begin{definitionbox}
\textbf{Definition (\textit{Agent-as-a-Judge}).}
An \textit{Agent-as-a-Judge} constructs a verdict about an evaluand through explicit checking operations and the use of their results.
At least one operation obtains evidence, performs a targeted verification, or reassesses a judgment using another check's findings or rationale, beyond replicating an overall score.
The resulting findings inform subsequent assessment or verdict formation.
We represent this process as:
\[
\begin{aligned}
a_t &\sim \pi(\cdot \mid s_t),  \qquad
o_t &= \Omega(a_t), \\
s_{t+1} &= u(s_t, a_t, o_t),  \qquad
\hat{y} &= g(s_T).
\end{aligned}
\]
Here $s_0$ encodes the evaluand $x$ and any initial context;
$s_t$ summarizes $\{(a_i,o_i)\}_{i<t}$ with that initialization;
$a_t$ is an evaluation action executed under policy $\pi$, with $o_t$ as its return;
and verdict $\hat{y}$ is produced when evaluation ends at $T$.
\end{definitionbox}
Concretely, $a_t$ may retrieve a document, probe workspace or environment state,
delegate a sub-check, exchange critiques with another judge, or retrieve a stored evaluation artifact;
$o_t$ is the return of that call---a retrieved passage, file contents, an environment status, or a sub-check result---
and $u$ incorporates it into $s_t$ for subsequent checks and the final verdict.
These evaluation processes range from structured workflows that coordinate predefined checks
to adaptive policies that revise what to inspect, how to verify it, and when to stop as evidence accumulates.
The policy $\pi$ can therefore prescribe a checking sequence or use intermediate observations to select subsequent actions;
likewise, $T$ can be fixed or determined during evaluation.
We distinguish prescribed execution, input-conditioned planning before execution, and observation-driven revision during execution.
These control choices apply across all four mechanisms and can coexist within a system.

\subsection{Survey Scope}
We include work whose main contribution is the evaluator and whose reported procedure satisfies the checking and result-use conditions above.
The operational boundary rests on what is checked, what result is produced, and where that result is used; the state--action formulation describes the process within this scope.
For tool-mediated evaluation, retrieved support or execution outcomes ground the verdict; in collaborative evaluation, targeted sub-checks or reassessment of peer rationales provide the corresponding findings.
Role names, additional calls, and voting alone do not establish this process.
Fixed retrieval or testing procedures can qualify, while generic task-execution agents do not qualify merely because they maintain state and act sequentially.
Appendix~\ref{app:inclusion} specifies the screening conditions and contrasting protocol examples.

\section{Methodology}\label{sec:methods}

Building the evaluation process formalized in Section~\ref{sec:background} involves organizing evaluative work, obtaining observations, retaining useful findings, and specifying the checks to perform.
We examine these design choices through four complementary mechanisms.
Multi-agent collaboration distributes checking tasks, information exchange, and verdict authority across evaluators (\S\ref{sec:multiagent});
tool integration provides interfaces for obtaining external observations and execution results (\S\ref{sec:tool});
memory governs which findings are stored, accessed, and updated (\S\ref{sec:memory});
and planning specifies checking targets, decompositions, and verification routes (\S\ref{sec:plan}).
These are composable design dimensions: a judge can coordinate multiple evaluators, equip them with tools, retain their findings in memory, and plan their checks within the same process.
Across these dimensions, we distinguish checks prescribed in advance, procedures tailored to the initial input, and checks revised from intermediate observations (Section~\ref{sec:background}).
Their interaction determines how intermediate results enter the evaluation state, support a verdict, and guide further checking; the synthesis returns to these connections through evidence integration and stopping decisions.
Figure~\ref{fig:taxonomy} places this methodological axis alongside Evaluation and Applications in the overall survey map.

\subsection{Multi-Agent Collaboration}
\label{sec:multiagent}
\begin{figure*}[t]
  \centering
  \includegraphics[width=\textwidth]{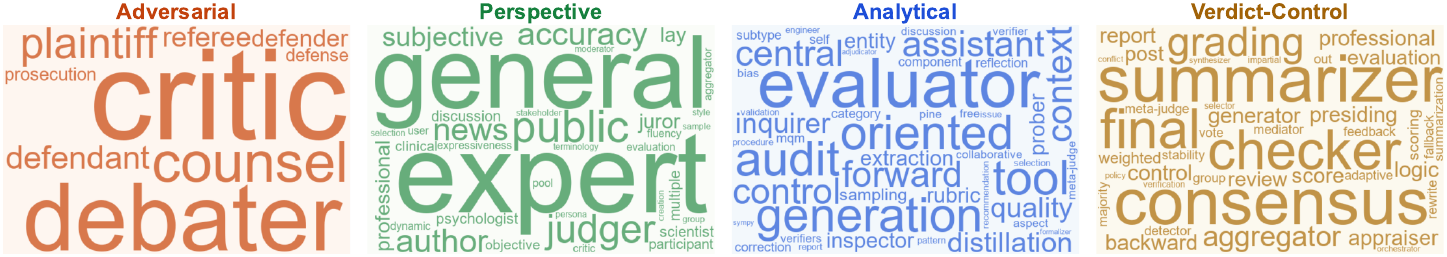}
  \caption{Role lexicon of subagents in multi-agent judges.
  The four panels group cleaned subagent names into adversarial, perspective, analytical, and verdict-control roles.}
  \label{fig:mad-role-wordclouds}
\end{figure*}

\begin{figure}[t]
  \centering
  \includegraphics[width=\linewidth]{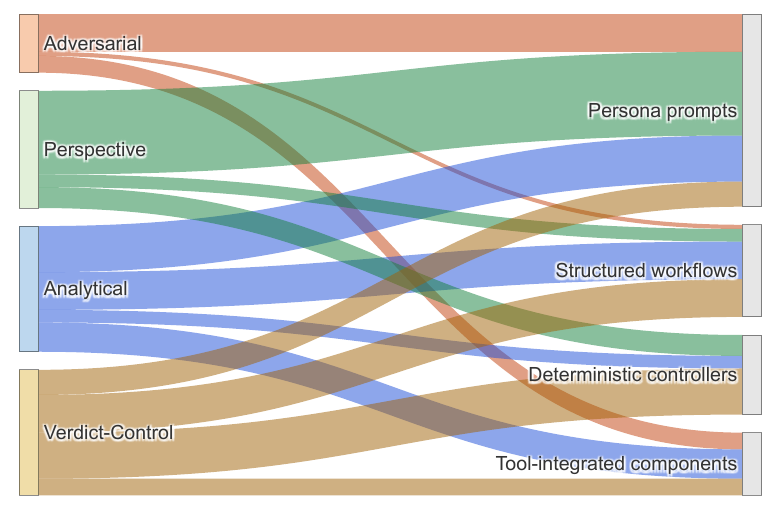}
  \caption{From functional role names to implementation classes.
  The Sankey maps subagent and component instances from functional roles to implementation classes.}
  \label{fig:mad-role-tech-sankey}
\end{figure}

Multi-agent collaboration distributes checking tasks and verdict authority across evaluators, making perspective, challenge, and resolution explicit parts of the judging protocol~\citep{chan2024ChatEval,Li2024MATEval,chen2025MultiAgentasJudge}.
Its central design choices concern who performs each check, which outputs are shared, and who can revise or resolve partial judgments.
We first identify those roles and their implementations, then trace how their outputs are exchanged and combined before the final verdict.
The inclusion criterion concerns the checking task and information flow: evaluators perform distinct verification tasks or use another check's findings or rationale in subsequent assessment.
Changing persona names while independently repeating the same overall scoring task does not qualify on its own.

\researchquestion{What Are the Subagents in a MAD Judge?}

\paragraph{Functionally} Reviewed subagent roles cluster into adversarial, perspective, analytical, and verdict-control categories (Figure~\ref{fig:mad-role-wordclouds}).
\emph{Adversarial} roles take a tentative answer or judgment and emit objections or cross-examinations that later stages must address.
AgentsCourt~\citep{he2024agentscourt} stages plaintiff and defendant arguments as advocacy transcripts that later feed legal decision-making.
\emph{Perspective} roles read the evaluand through different identities or expertise and return alternative rationales for the same case.
AgentsBench~\citep{jiang2025agentsbench} pairs professional and lay judge agents whose independent sentencing recommendations feed later deliberation.
\emph{Analytical} roles transform the evaluand or an upstream answer into inspectable intermediate evidence.
GEMA-Score~\citep{zhenxuan2025GEMA-Score} extracts clinical entities that downstream scoring must cite, while MATH-VF~\citep{MATH-VF} formalizes informal reasoning into SimpleMath statements that critics and tool agents verify step by step.
\emph{Verdict-control} roles take partial judgments, discussion state, or protocol signals as input to aggregate, stop, reroute, or authorize a verdict.
MATEval~\citep{Li2024MATEval} routes discussion through feedback and summarization before the final verdict is issued, while CollabEval~\citep{gupta2026CollabEval} applies consensus and unchanged-result checks to decide whether deliberation should continue or close.

\paragraph{Technically} The same role labels map to different implementation classes (Figure~\ref{fig:mad-role-tech-sankey}).
A named subagent may be a persona prompt, a structured workflow, a deterministic controller, or a tool-integrated component, and the label alone does not specify which control mode applies.
\emph{Persona prompts} specify an evaluator's perspective or checking remit within a protocol.
Their contribution depends on the surrounding process: debate systems expose peer rationales for further assessment~\citep{chan2024ChatEval,gupta2026CollabEval}, whereas Multi-Agent Verification~\citep{lifshitz2025multi} assigns distinct verification aspects and aggregates their approvals without peer deliberation.
\emph{Structured workflows} chain artifact-producing steps: an extractor, formalizer, critic, or final judge consumes a specific upstream output and emits a constrained downstream artifact.
\emph{Deterministic controllers} apply agreement ratios, unchanged-result checks, or stability tests to decide whether deliberation should stop, continue, or escalate~\citep{gupta2026CollabEval,hu2025AdaptiveStability}.
\emph{Tool-integrated components} operationalize verification through predefined interfaces, as when AgentV-RL~\citep{AgentV-RL} embeds Python calls inside forward and backward verifier trajectories.
Adversarial and perspective names mostly resolve to persona prompts, while verdict-control names more often split across deterministic controllers and structured workflows.
The implementation class identifies how a role is realized; checking responsibilities, information dependencies, and continuation rules identify its place in the evaluation process.

\researchquestion{How Do Subagents Collaborate Within a MAD Judge?}
Subagent roles identify who participates in judgment production, but the final verdict depends on how those roles connect their partial outputs before closure.
Reviewed MAD judges coordinate those connections through sequential staging, reciprocal deliberation, and parallel aggregation.

\paragraph{Sequential}
Sequential collaboration makes later stages depend on artifacts produced upstream.
In evidence-to-judgment pipelines, advocacy transcripts or collected software evidence become inputs for later judgment: AgentsCourt~\citep{he2024agentscourt} feeds plaintiff and defendant arguments into legal decision-making, while CodeVisionary~\citep{wang2025CodeVisionary} routes software requirements through runtime, static-analysis, UI, and web evidence before scoring.
In rubric and decomposition pipelines, the upstream artifact constrains what downstream judges can score.
Upstream artifacts include refined rubrics, extracted clinical entities, formalized statements, and proposed robot-event timelines that downstream agents score or verify~\citep{xie2024grade,zhenxuan2025GEMA-Score,MATH-VF,hu2026ARS}.
The benefit is traceability: the verdict can be followed through intermediate artifacts.
The risk is upstream lock-in, because a flawed rubric, extraction, formalization, or evidence localization constrains every later judgment.

\paragraph{Reciprocal}
Reciprocal collaboration turns another subagent's statement into new input for revision.
The relevant operation is state change after peer exposure: a subagent reads another rationale, updates its own position, and the protocol tests whether disagreement has narrowed enough for closure.
AgentsBench~\citep{jiang2025agentsbench} begins with independent sentencing recommendations, then lets professional and lay judge agents deliberate until a presiding judge detects convergence or resolves residual disagreement.
CollabEval~\citep{gupta2026CollabEval} runs initial evaluation, discussion, and final-judge phases, with consensus and unchanged-result checks deciding whether deliberation has stabilized.
HiMATE~\citep{zhang2025HiMATE} uses tiered agree/disagree turns to resolve translation-evaluation ambiguity, while Efficient LLM Safety Evaluation~\citep{wu2025EfficientSafetyMAD} uses critic--defender exchanges to surface safety risks before a judge assigns the risk score.
This form is useful when objection and revision can expose hidden assumptions, but its closure rule must guard against agreement that suppresses useful minority evidence.

\paragraph{Parallel}
Parallel collaboration preserves independent assessments until a resolver combines them.
The resolver determines what independence contributes to the verdict.
Formulaic merge rules make aggregation reproducible, as when AgentV-RL~\citep{AgentV-RL} combines forward and backward verifier confidence or Multi-Agent Verification~\citep{lifshitz2025multi} sums independent aspect-level approvals.
Weighted or threshold-based resolution governs whether partial assessments authorize a verdict, including meta-judgments and supporting/refuting arguments whose confidence gap controls further evidence gathering~\citep{braun2025MetaJudges,li2026DARE}.
No-human-in-the-loop~\citep{No-human-in-the-loop} synthesizes labels from many LLM configurations through majority vote and a severity hierarchy for persistent conflicts.
Parallel systems differ in what the resolver makes visible: a reproducible formula, an authority point, or a fallback path for unresolved disagreement.

\subsection{Tool Integration}
\label{sec:tool}
\begin{figure}[t]
  \centering
  \includegraphics[width=0.95\linewidth]{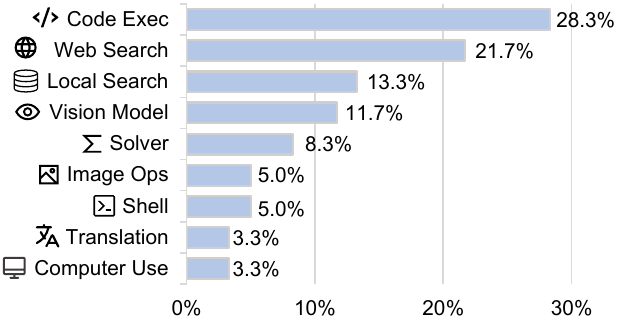}
  \caption{Distribution of 60 tool-interface assignments in a manually coded subset of 34 tool-augmented judges.
  Each bar shows one interface's share of the total; papers using multiple interfaces contribute one count per interface type.}
  \label{fig:tool-overview}
\end{figure}

Tool integration allows an \textit{Agent-as-a-Judge} to move beyond fixed-context inference by actively obtaining information or signals outside the initial prompt.
A prompt-only judge must infer quality, correctness, or risk from the evaluand and the context already placed in the prompt.
A tool-augmented judge can instead retrieve external facts, execute programmatic checks, inspect multimodal or interactive artifacts, and use the resulting observations to ground subsequent judgment.
We therefore analyze tool integration along three axes: interfaces, selection, and function---which observations are accessible, when an interface is invoked, and how its results support evaluation.

\researchquestion{What Interfaces Are Used?}
Figure~\ref{fig:tool-overview} shows which tool interfaces appear most often in the manually coded subset.
Execution and search interfaces dominate.
Code execution supports programmatic checks~\citep{xu2025TIRJudge,AgentV-RL}, while web search~\citep{OpenReward,TALE} and local search~\citep{auto-eval-judge,TATTOO} bring external or stored evidence into the judging context.
Vision models~\citep{TIM-PRM} and image operations~\citep{ding2025ARMThinker,wang-etal-2025-unified} support multimodal inspection.
Solvers~\citep{MATH-VF} and shell access~\citep{wang2025CodeVisionary} serve more specialized verification or runtime-inspection needs.
Translation~\citep{ahmad-etal-2025-urdufactcheck,li2024toolaugmented} and computer-use interfaces~\citep{wang2025CodeVisionary,VAGEN} remain less common, but they mark cases where judges translate queries or evidence across languages or act in interactive environments.

\begin{figure*}[t]
  \centering
  \includegraphics[width=1\linewidth]{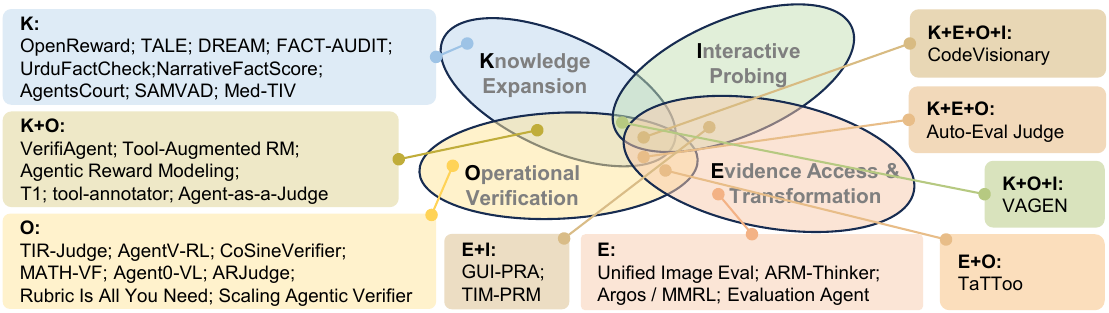}
  \caption{Functional view of tool use in \textit{Agent-as-a-Judge}.
  Four overlapping regions show that judges can chain multiple tool roles within one evaluation.}
  \label{fig:tool-functions}
\end{figure*}

\researchquestion{How Are Tools Selected?}
Tool selection determines when a judge should call a tool, which tool it should call, and whether the returned result is enough for judgment.
Across existing systems, this control becomes progressively more adaptive: the judge may follow fixed rules, call tools when current evidence is weak, use observations from one tool call to guide later calls, or actively create new observations.
1) Fixed-rule selection uses predefined signals: the modality, domain tag, or answer type decides whether the case goes to search, code execution, or a solver.
VerifiAgent~\citep{han2025VerifiAgent} routes checks by reasoning type, while Auto-Eval Judge~\citep{auto-eval-judge} dispatches cases according to verification type.
2) Evidence-triggered selection asks whether the current evidence is enough.
The judge keeps prompt-only reasoning as the default, but escalates to retrieval when the current trace lacks support, as in OpenReward~\citep{OpenReward} and TALE~\citep{TALE}; UrduFactCheck~\citep{ahmad-etal-2025-urdufactcheck} applies the same idea in multilingual fact checking.
3) Observation-guided selection lets the result of one tool call shape the next tool call or checking step.
This pattern is stronger than one-shot escalation because tool outputs become part of the control signal for later checks.
Accumulated findings guide further perception steps, cross-channel inspections, and targeted state, trace, or temporal-window queries~\citep{GUI-PRA,wang2025CodeVisionary,sun2026VeriCUA,sanayei2026Sauron,hu2026ARS}.
4) Active probing runs a new test or acts on the environment to surface observations that were not available in the initial context.
Environment interaction and executable tests expose additional evidence for GUI-task completion, code correctness, and application behavior~\citep{VAGEN,agentic-verifier,chi2026InteractionJudge}.

\researchquestion{What Functions Do Tools Serve?}
Tool use gives the judge capacities that are unavailable from the prompt and the candidate answer alone (Figure~\ref{fig:tool-functions}).
After a tool is called, the observations it returns can support judgment in several ways.
They may add missing knowledge, expose inspectable evidence, test correctness operationally, or make the final score depend on an interaction trace.
These uses form four recurring roles.
1) Knowledge expansion links the verdict to information outside parametric memory so judgments can depend on up-to-date or domain-specific facts.
OpenReward~\citep{OpenReward} and TALE~\citep{TALE} retrieve or escalate to outside facts when parametric knowledge is insufficient, while DREAM~\citep{DREAM} derives case-specific checks for deep research evaluation.
Domain stores and structured pipelines query narrative graphs~\citep{jeong-etal-2025-agent}, legal document collections~\citep{he2024agentscourt,devadiga2025samvad}, or medical evidence bases~\citep{Med-TIV} before scoring.
FACT-AUDIT~\citep{lin-etal-2025-fact} extends the same grounding pattern to factual audit, while UrduFactCheck~\citep{ahmad-etal-2025-urdufactcheck} applies it in multilingual fact checking.
2) Evidence access and transformation converts opaque evaluands into localized, inspectable artifacts that the judge can cite rather than summarize.
Auto-Eval Judge~\citep{auto-eval-judge} exposes support through log search, while TaTToo~\citep{TATTOO} does so through table querying.
In multimodal settings, OCR, visual differencing, cropping, and specialized perception tools localize textual, visual, or spatial evidence for inspection~\citep{MMRL,wang-etal-2025-unified,ding2025ARMThinker,tan2026SpatialCritic}.
The verdict therefore rests on citable support rather than on an overall impression.
3) Operational verification uses executable interpreters, solvers, and programmatic checkers to test intermediate steps, format constraints, or derived consequences, turning plausibility into a failed check, a counterexample, or a satisfied constraint.
Existing systems automate code-oriented checking~\citep{xu2025TIRJudge}, reasoning-type verification~\citep{han2025VerifiAgent}, and math-oriented formal checking~\citep{MATH-VF}.
The same pattern also supports training-time scoring~\citep{AgentV-RL}, tool-integrated answer checking~\citep{T1}, and format-oriented checking~\citep{xu2025ARJudge}.
4) Trace-based judgment applies when the initial context cannot settle the verdict and the score depends on the trace produced by tool-mediated interaction.
Some judges question visual reasoning steps~\citep{TIM-PRM} or test the evaluand directly~\citep{agentic-verifier}; others act on GUI history~\citep{GUI-PRA} or runtime state~\citep{VAGEN}.
A few chain repository inspection with runtime, UI, and web probes in one evaluation~\citep{wang2025CodeVisionary}.
Auxiliary LLM modules or sub-agents route cases~\citep{peng2025AgenticRM,auto-eval-judge}, extract evidence, or aggregate observations within the loops above~\citep{wang2025CodeVisionary}.


\subsection{Memory}
\label{sec:memory}
\begin{figure*}[t]
  \centering
  \includegraphics[width=\textwidth]{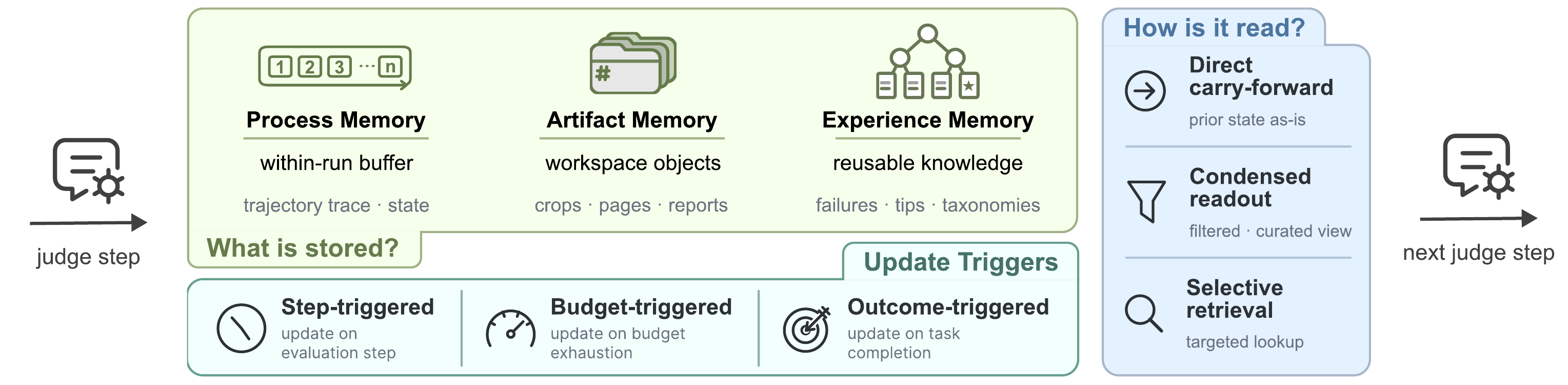}
  \caption{Memory in \textit{Agent-as-a-Judge} can be analyzed as a write--manage--read loop over evaluation state.
  The stored state may take the form of process memory, artifact memory, or experience memory.
  Later judgments either receive this state directly, access a filtered or summarized form, or target specific artifacts, failures, or templates through retrieval.}
  \label{fig:memory-mechanisms}
\end{figure*}

Memory allows an \textit{Agent-as-a-Judge} to carry evaluation state across steps, cases, or interactions rather than treating each judgment as an isolated inference.
A memory mechanism makes the lifecycle of reusable findings explicit through storage, access, and update operations.
We therefore analyze memory along a write--manage--read loop (Figure~\ref{fig:memory-mechanisms}) through three questions: what evaluation state is preserved, how selectively it is read back, and what event authorizes an update.
The same system may combine several patterns across these axes.

\researchquestion{What Memory Is Stored?}
Three storage structures recur across agentic judges, although a single system may combine them.
1) \emph{Process memory} accumulates evaluation state step by step within a run rather than as a pre-indexed collection of objects.
Later steps read this buffer back to change an ongoing evaluation, most often in one of two ways.
For \emph{verification}, the buffer holds proof state from search or checking---queries, evidence summaries, reflections, and support or contradiction signals---that later steps consume to refine evidence and determine a verdict.
Fact-checking judges such as TALE~\citep{TALE} exemplify this pattern through iterative query--evidence--reflection buffers.
For \emph{interaction scoring}, the buffer keeps enough of a long interaction or dialogue to score the current step without forwarding the entire transcript unchanged.
Interactive and dialogue judges maintain bounded history through relevance filtering, progressive summarization, or evolving internal state rather than unbounded transcript growth~\citep{GUI-PRA,zhang2025sentient}.
2) \emph{Artifact memory} retains case-specific artifacts or reports that later checking steps reopen by index or retrieval rather than as undifferentiated context.
Multimodal and code-inspection judges index crops, document pages, or stage-wise analysis reports for later tool calls and evidence-grounded scoring~\citep{ding2025ARMThinker,wang2025CodeVisionary}.
Agent-as-a-Judge~\citep{zhuge2025AgentasaJudge} exposes similar modules, but ablation evidence suggests limited benefit and possible error propagation, making it a cautionary rather than an anchor example.
3) \emph{Experience memory} curates prior evaluations into structured knowledge---failure pools, scenario taxonomies, or critique templates---that later cases draw on rather than re-deriving from scratch~\citep{lin-etal-2025-fact,yu2025TableCritic}.

\researchquestion{How Is Memory Read?}
Read access ranges from direct carry-forward to condensed readout to selective retrieval.
1) \emph{Direct carry-forward} places stored state into the next judging context without an explicit choice among memory items.
Verification-oriented process memory~\citep{TALE} and interaction-scoring buffers~\citep{zhang2025sentient} follow this pattern cleanly.
Multi-agent deliberation systems also inherit full discussion records~\citep{Li2024MATEval,chen2025MultiAgentasJudge,jiang2025agentsbench}, but transcript availability alone does not establish selective access, compression, or curation of memory.
2) \emph{Condensed readout} first constructs a smaller usable view through filtering or summarization, then supplies that view to the judge.
Long-horizon GUI evaluation depends on this pattern when current-step reward scoring must use a compressed interaction history rather than the full buffer~\citep{GUI-PRA}.
3) \emph{Selective retrieval} targets a particular stored object when the current step needs it---by index lookup, pool search, or template routing.
Strongest cases index specific multimodal artifacts or document pages~\citep{ding2025ARMThinker}, draw failure cases from audit pools~\citep{lin-etal-2025-fact}, or route critique generation through an error-type template tree~\citep{yu2025TableCritic}.
CodeVisionary~\citep{wang2025CodeVisionary} is a weaker fit because later stages clearly reuse distilled reports, but the available evidence is less explicit about retrieval among report items.

\researchquestion{How Is Memory Updated?}
The same system may combine several update triggers within one evaluation run.
1) \emph{Step-triggered} changes memory after each evaluation step, tool observation, reflection, report, or discussion round---the write cadence follows the evaluation process itself.
Iterative fact checking extends process memory on each query--evidence cycle~\citep{TALE}; deliberation judges append round-level discussion state~\citep{Li2024MATEval,chen2025MultiAgentasJudge,jiang2025agentsbench}; and multimodal inspection adds tool-produced artifacts to artifact memory as checking proceeds~\citep{ding2025ARMThinker,wang2025CodeVisionary}.
2) \emph{Budget-triggered} changes memory because the working state has become too long, too noisy, or too stale to use directly.
GUI interaction scoring filters recent history and summarizes older steps once interaction length grows~\citep{GUI-PRA}, while dialogue judging updates a compact running state each turn rather than accumulating an unbounded transcript~\citep{zhang2025sentient}.
3) \emph{Outcome-triggered} changes memory only after an evaluation outcome identifies something worth preserving as future guidance---a low-score audit case, failure pattern, successful critique, or new error type.
Within-target factual audit accumulates weak cases and revises scenario taxonomies after poor outcomes~\citep{lin-etal-2025-fact}, while table critique expands its template tree after successful refinements across instances~\citep{yu2025TableCritic}.
Deliberation systems that only record round-by-round discussion without budget-triggered or outcome-triggered updates remain at the weaker end of this axis, closer to deliberation carryover than to the managed patterns above.

\subsection{Planning}
\label{sec:plan}
\begin{figure}[t]
  \centering
  \includegraphics[width=1\linewidth]{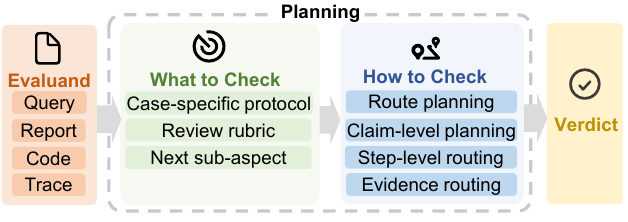}
  \caption{Planning in \textit{Agent-as-a-Judge} systems can externalize either what to check or how to gather proof before the verdict.}
  \label{fig:planning-pipeline}
\end{figure}

Planning makes the checking procedure explicit by specifying what should be inspected and how proof should be gathered.
A plan can organize a case-specific workflow before execution or guide successive checks as observations accumulate.
The planning object is the checking procedure: its targets, decomposition, and route through the available evidence sources.
We therefore distinguish two forms: what-to-check planning, which defines the target of evaluation, and how-to-check planning, which defines the route of evidence gathering (Figure~\ref{fig:planning-pipeline}).

\researchquestion{What to Check?}
What-to-check planning defines the target that subsequent requirement checks must address, potentially after an initial inspection of the available context.
The planned target can take three forms.
1) Case-specific protocols turn the current query or evaluand into checks tailored to that case.
DREAM~\citep{DREAM}, for example, derives report-specific fact checks and validation questions before executing the evaluation protocol.
2) Rubric planning searches or constructs the criteria that will guide later judgment.
Criteria can be selected through rubric search or constructed from the task and inspected environment context~\citep{apres2026,sun2026VeriCUA}.
3) Sub-aspect planning chooses which aspect of an open-ended evaluand should be inspected next.
Evaluation Agent~\citep{zhang2025EvaluationAgent} uses this pattern in visual evaluation by moving across proposed sub-aspects over successive rounds.


\researchquestion{How to Check?}
How-to-check planning defines the route by which evidence is gathered once the evaluation target is fixed.
This route can specify the order of checks, the granularity of intermediate steps, the evidence source for each step, or the condition for stopping.
1) Route planning writes an evaluation plan before scoring and then follows that route to the verdict, as in EvalPlanner~\citep{evalplanner2025}.
2) Claim-level planning decomposes a candidate solution into smaller claims and verifies them along an explicit path.
AgentV-RL~\citep{AgentV-RL}, for example, verifies atomic claims through forward or backward verifier trajectories.
3) Step-level routing decides what kind of evidence each step requires before invoking a tool.
TIM-PRM~\citep{TIM-PRM} routes steps toward visual, knowledge-based, or logical checks, while VerifiAgent~\citep{han2025VerifiAgent} checks completeness and consistency before selecting the matching tool.
4) Evidence-channel dispatch sends a case to the evidence source or evaluation handler required by the planned procedure.
CodeVisionary~\citep{wang2025CodeVisionary} dispatches requirements across runtime, static-analysis, UI, and web evidence channels, while Auto-Eval Judge~\citep{auto-eval-judge} routes cases according to verification type and expands retrieval when proof is thin.

\begin{table*}[t]
  \centering
  \caption{Planning mechanisms in \textit{Agent-as-a-Judge}.
  What-to-check planning decides the evaluation target, while how-to-check planning decides the evidence-gathering route.}
  \label{tab:planning-summary}
  \scriptsize
  \setlength{\tabcolsep}{4.5pt}
  \renewcommand{\arraystretch}{1.24}
  \begin{tabular}{@{}
    >{\raggedright\arraybackslash}p{2.25cm}
    >{\raggedright\arraybackslash}p{3.7cm}
    >{\raggedright\arraybackslash}p{2.9cm}
    >{\raggedright\arraybackslash}p{7.6cm}
  @{}}
    \toprule
    \multicolumn{1}{l}{\textbf{System}} &
    \multicolumn{1}{l}{\textbf{Planned element}} &
    \multicolumn{1}{l}{\textbf{Decision point}} &
    \multicolumn{1}{l}{\textbf{Mechanism}} \\
    \midrule
    \multicolumn{4}{@{}c@{}}{\colorbox{softsage!30}{\makebox[\dimexpr\textwidth-2\fboxsep\relax][c]{\strut\textbf{What-to-check planning}}}} \\
    \addlinespace[2pt]
    DREAM~\citep{DREAM} & Query-specific evaluation protocol & Before protocol execution & Turns the query into report-specific fact checks and validation questions. \\
    \addlinespace[1pt]
    APRES~\citep{apres2026} & Review rubric & During rubric search & Searches over candidate criteria and keeps the rubric used for revision. \\
    \addlinespace[1pt]
    Evaluation Agent~\citep{zhang2025EvaluationAgent} & Next sub-aspect & Across proposal rounds & Moves from one aspect to the next and stops when the accumulated evidence is sufficient. \\
    \midrule
    \multicolumn{4}{@{}c@{}}{\colorbox{softsky!30}{\makebox[\dimexpr\textwidth-2\fboxsep\relax][c]{\strut\textbf{How-to-check planning}}}} \\
    \addlinespace[2pt]
    EvalPlanner~\citep{evalplanner2025} & Unconstrained evaluation plan & Before scoring & Writes the checking route first and then executes it. \\
    \addlinespace[1pt]
    AgentV-RL~\citep{AgentV-RL} & Verification path over atomic claims & Per candidate solution & Splits a solution into atomic claims and verifies them forward or backward. \\
    \addlinespace[1pt]
    TIM-PRM~\citep{TIM-PRM} & Per-step evidence route & Before each tool call & Decides whether the next check needs vision, outside knowledge, or logic. \\
    \addlinespace[1pt]
    VerifiAgent~\citep{han2025VerifiAgent} & Two-layer verification policy & Before tool invocation & Checks completeness and consistency before selecting the matching tool. \\
    \addlinespace[1pt]
    CodeVisionary~\citep{wang2025CodeVisionary} & Requirement-level evaluation plan & Before context collection & Gathers runtime, static-analysis, UI, and web evidence in the planned order. \\
    \addlinespace[1pt]
    Auto-Eval Judge~\citep{auto-eval-judge} & Verification-type dispatch policy & After checklist generation & Routes cases by verification type and expands retrieval when proof is thin. \\
    \bottomrule
  \end{tabular}
\end{table*}

\subsection{Synthesis}

\begin{figure*}[t]
  \centering
  \includegraphics[width=\textwidth]{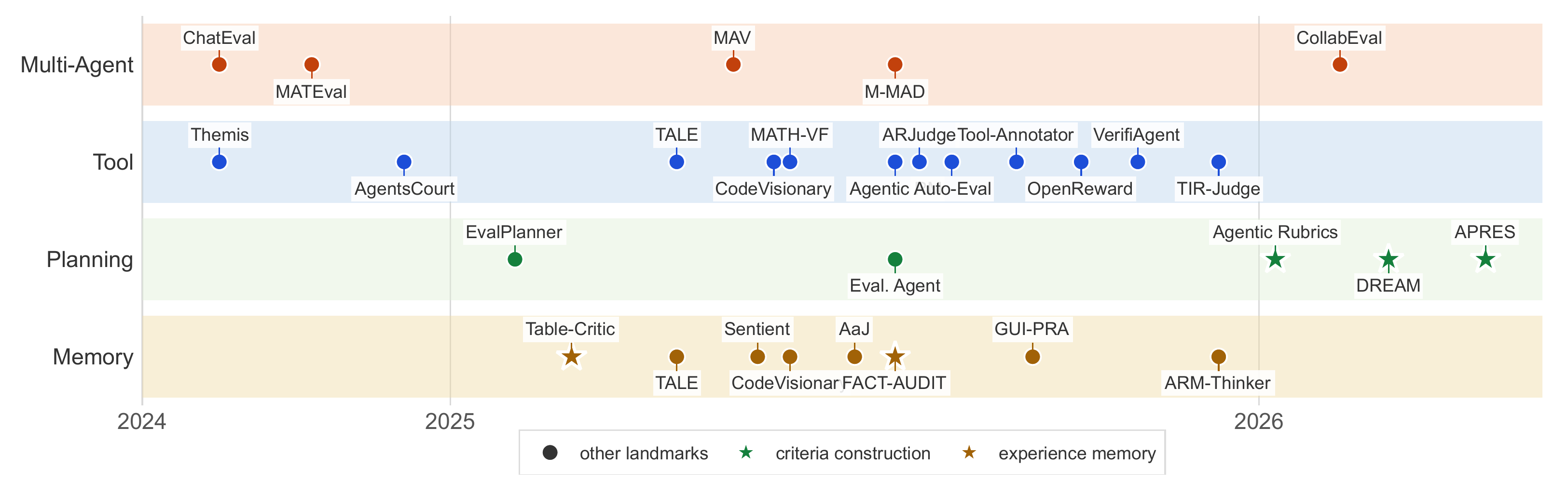}
  \caption{Representative adoption timeline of the four methodological mechanisms in \textit{Agent-as-a-Judge}.
  A method may appear in more than one lane when it instantiates multiple mechanisms.
  Stars mark two sparse patterns discussed in the text:
  rubric construction on Planning and experience memory on Memory.}
  \label{fig:mechanism-timeline}
\end{figure*}

The four mechanisms offer complementary views of how evaluation processes are built and how their intermediate results support judgment.
We first arrange representative systems on a calendar timeline stratified by mechanism (Figure~\ref{fig:mechanism-timeline}).
Multi-agent debate that diversifies and challenges single-model judgment is a comparatively early---and still active---direction for judges~\citep{chan2024ChatEval,Li2024MATEval}.
Tool augmentation appears on a similar early timeline, yet the toolkit has kept shifting:
calculator-style and mixed external APIs opened the loop~\citep{li2024toolaugmented},
while recent designs concentrate on code execution and search as the dominant evidence stack~\citep{OpenReward,TALE,xu2025TIRJudge,auto-eval-judge}.
More recently, as rubric-based evaluation has become a focal theme for LLM-based judgment~\citep{chen2026rubrics},
planning attention has moved toward constructing atomic, case-specific, and independently verifiable criteria before scoring~\citep{apres2026,DREAM,Agentic-Rubrics}.
Memory has long been a dedicated research focus in general LLM-based agents for long-horizon interaction and self-evolution~\citep{zhang2025agentmemory,du2026agentmemory},
where reusable experience across sessions and tasks is widely treated as central to sustained performance.
In Agent-as-a-Judge, by contrast, most systems keep only a short within-run buffer of evaluation state~\citep{TALE,GUI-PRA},
and few curate reusable experience across cases~\citep{lin-etal-2025-fact,yu2025TableCritic}.

\paragraph{From findings to verdicts}
Across these designs, evidence integration connects checking operations to final judgment.
Tools and sub-judges produce observations or partial assessments, memory retains them for reuse, and the checking plan organizes which claims they address.
The update $u$ incorporates these results into evaluation state, while $g$ resolves that state into a verdict.
Formulaic aggregation combines partial scores or approvals, as in forward/backward verifier combination and independent aspect verification~\citep{AgentV-RL,lifshitz2025multi}.
Deliberative resolution instead lets evaluators examine competing rationales before a presiding or final judge closes the case~\citep{jiang2025agentsbench,gupta2026CollabEval}.
Resolution rules can handle conflicting labels through severity hierarchies or require source-grounded support before accepting a criterion verdict~\citep{No-human-in-the-loop,sanayei2026Sauron,sun2026VeriCUA}.
These choices determine how supporting, conflicting, or differently scoped findings contribute to the verdict.

\paragraph{From findings to further checks}
The same intermediate results can determine whether evaluation should continue.
Prescribed workflows reach the verdict after their planned checks or debate rounds~\citep{evalplanner2025,chan2024ChatEval}; observation-driven policies use returned findings to revise checks or determine the stopping point $T$.
Consensus, unchanged-result checks, and adaptive stability detection provide explicit closure signals in multi-agent deliberation~\citep{gupta2026CollabEval,hu2025AdaptiveStability}.
Evidence-seeking judges connect continuation to unresolved support through retrieval escalation, sub-aspect inspection, or renewed exploration of supporting and refuting evidence~\citep{OpenReward,TALE,auto-eval-judge,zhang2025EvaluationAgent,li2026DARE}.
Here planning specifies the next verification target or route, tool selection supplies the relevant observation interface, and coordination assigns the check and its resolution authority.
Agreement, stability, and evidence sufficiency thus serve distinct control signals; their use does not by itself establish calibrated confidence.
Together, evidence integration and continuation policies explain how the four mechanisms turn accumulated findings into a verdict or a reason to investigate further.


\section{Evaluation}\label{sec:evaluation}

\definecolor{fnMapRMPattern}{HTML}{4D84C4}

\begin{figure*}[t]
  \centering
  \includegraphics[width=0.85\textwidth]{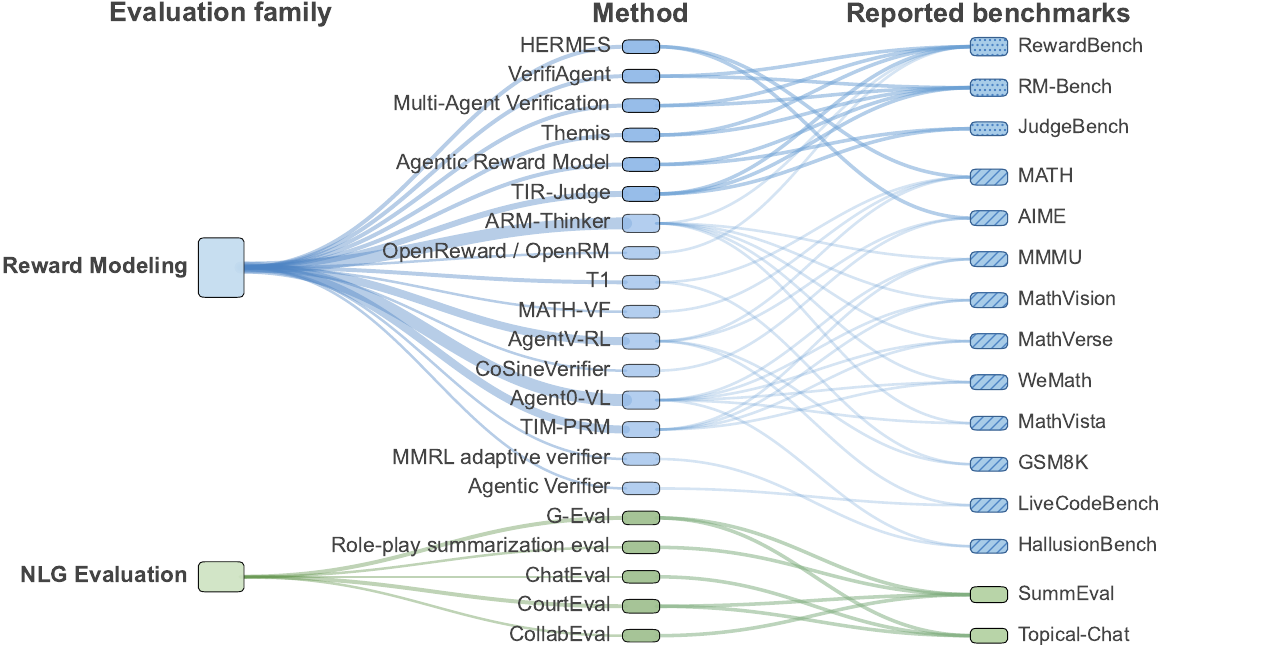}
  \caption{Evaluation--benchmark coverage map.
  Links indicate that a method reports results on, or is explicitly evaluated on, a reported benchmark.
  Right-node fill and pattern encode benchmark artifact type:
  \textcolor{fnMapRMPattern}{$\bullet\bullet\bullet$}~fixed response or judgment sets packaged for direct comparison;
  \textcolor{fnMapRMPattern}{/\kern0.12em/\kern0.12em/}~task problems whose reported results require candidate generation before judging;
  and green nodes, NLG evaluation datasets.}
  \label{fig:evaluation-benchmark-map}
\end{figure*}

Evaluation protocols have not kept pace with method development (Section~\ref{sec:methods}): most reported evidence still concentrates on shared judge-quality and task benchmarks (Figure~\ref{fig:evaluation-benchmark-map}).
We organize this evidence by the signal each system is validated to produce.
\textbf{Reward modeling} turns responses, trajectories, or reasoning traces into preference, reliability, or correctness signals~\citep{christiano2017deep,stiennon2020learning,ouyang2022training,cobbe2021training,lightman2023lets}.
\textbf{NLG evaluation} instead scores summaries, conversations, stories, and other generated texts along subjective or multidimensional quality axes~\citep{Zheng2023Judging,liu2023G-Eval,li2025LLM-as-a-Judge-Survey}.

\subsection{Reward Modeling}
\label{sec:evaluation-rm}

Reward-model signals support candidate ranking, reward assignment, data filtering, and reasoning verification.
Recent agentic variants change the judging procedure through tool-integrated reasoning~\citep{xu2025TIRJudge,li2024toolaugmented}, planner--coder decomposition~\citep{peng2025AgenticRM}, multi-verifier aggregation~\citep{lifshitz2025multi}, or verification-oriented scoring~\citep{han2025VerifiAgent}.

Reward models are evaluated in three settings.
The first is \emph{test-time selection}: benchmarks such as MATH~\citep{MATH} and AIME~\citep{AIME} supply problems and references, and the judge selects, accepts, or ranks generated candidates.
For example, AgentV-RL~\citep{AgentV-RL} distills an agentic verifier and measures Best-of-$N$ selection and verifier-guided actor revision.
Results therefore depend on the candidate-generation pool as well as the judge.
The second is \emph{downstream training}, which measures the reward signal's effect on policy learning, data construction, or actor revision.
Reported uses include PPO/RLTAF, DPO-data construction, interaction-grounded rewards for application generation, and selection of robot action chunks for policy learning~\citep{li2024toolaugmented,peng2025AgenticRM,chi2026InteractionJudge,hu2026ARS}.
The third is a \emph{benchmark} setting using fixed judgment artifacts for cross-model comparison.
Pairwise suites such as RewardBench~\citep{lambert2024rewardbench}, RM-Bench~\citep{liu2024rmbench}, and JudgeBench~\citep{tan2025judgebench} dominate this surface; VerifyBench~\citep{yan2025verifybench}, ProcessBench~\citep{zheng2025processbench}, and IFBench~\citep{peng2025AgenticRM} extend it to reference-grounded verification, step-level error localization, and instruction-following constraints, respectively.
Table~\ref{tab:evaluation-rm-benchmarks} catalogs evaluand form, domain coverage, and judgment signal across these benchmarks.
These suites measure answer selection, reward reliability, or reference agreement~\citep{lambert2024rewardbench,liu2024rmbench,tan2025judgebench}, but do not isolate the contributions of planning, tool use, evidence acquisition, aggregation, or stopping.
High scores on these suites also may fail to predict gains in test-time selection or downstream training~\citep{liu2024rmbench,viswanathan2026checklists}.
Environment-aware evaluation complements fixed-artifact suites by testing trajectory verification through interaction or inspection of post-execution state~\citep{shi2026AJBench,sun2026VeriCUA}; its results should distinguish evidence-access conditions and report decision coverage alongside accuracy when abstention is allowed.

\begin{table*}[t]
    \centering
    \caption{Reward-model and verifier benchmarks.}
    \label{tab:evaluation-rm-benchmarks}
    \scriptsize
    \setlength{\tabcolsep}{2.0pt}
    \renewcommand{\arraystretch}{1.05}
    \begin{tabular*}{\textwidth}{@{\extracolsep{\fill}} >{\raggedright\arraybackslash}p{0.22\textwidth} >{\raggedright\arraybackslash}p{0.11\textwidth} >{\raggedright\arraybackslash}p{0.12\textwidth} >{\raggedright\arraybackslash}p{0.20\textwidth} >{\raggedright\arraybackslash}p{0.17\textwidth} >{\raggedright\arraybackslash}p{0.08\textwidth} C{0.05\textwidth} @{}}
      \toprule
      \textbf{Benchmark / dataset} & \textbf{Evaluand form} & \textbf{Evaluand} & \textbf{Domain} & \textbf{Judgment signal} & \textbf{Scale} & \textbf{URL} \\
      \midrule
      RewardBench~\citep{lambert2024rewardbench} & pair-wise & response pair & chat, safety, reasoning, code/math & chosen--rejected preference & 2,985 & \githubartifact{https://github.com/allenai/reward-bench}\,\hfdataset{https://hf.co/datasets/allenai/reward-bench} \\
      RM-Bench~\citep{liu2024rmbench} & pair-wise & response pair & chat, code, math, safety & subtle correctness under style variation & 1,237 & \githubartifact{https://github.com/THU-KEG/RM-Bench}\,\hfdataset{https://hf.co/datasets/THU-KEG/RM-Bench} \\
      JudgeBench~\citep{tan2025judgebench} & pair-wise & response pair & knowledge, reasoning, math, code & objective correctness label & 620 & \githubartifact{https://github.com/ScalerLab/JudgeBench}\,\hfdataset{https://hf.co/datasets/ScalerLab/JudgeBench} \\
      IFBench~\citep{peng2025AgenticRM} & pair-wise & response pair & instruction following & constraint satisfaction & 444 & \githubartifact{https://github.com/THU-KEG/Agentic-Reward-Modeling}\,\hfdataset{https://hf.co/datasets/THU-KEG/IFBench} \\
      VerifyBench~\citep{yan2025verifybench} & point-wise & completion & reference-based answer verification & binary correctness against reference & 3,000 & \webartifact{https://zju-real.github.io/VerifyBench/}\,\hfdataset{https://hf.co/datasets/ZJU-REAL/VerifyBench} \\
      ProcessBench~\citep{zheng2025processbench} & step-wise & reasoning trace & mathematical reasoning & earliest-error step label & 3,400 & \githubartifact{https://github.com/QwenLM/ProcessBench}\,\hfdataset{https://hf.co/datasets/Qwen/ProcessBench} \\
      \bottomrule
    \end{tabular*}
    \vspace{2pt}
    \parbox{\textwidth}{\scriptsize
      \textit{Note:} Evaluand-form, evaluand, domain, and judgment-signal cells use controlled labels.
      Pair-wise rows provide candidate responses or completions with labels, whereas point-wise and step-wise rows test reference-based verification or process-error localization.
      In the URL column, \faIcon{github} marks a GitHub artifact, \faIcon{database} marks a Hugging Face dataset artifact, and \faIcon{link} marks a project page.}
  \end{table*}

\begin{table*}[t]
  \centering
  \caption{Local reward-model results on shared judge-quality benchmarks.}
  \label{tab:evaluation-rm-results}
  \scriptsize
  \setlength{\tabcolsep}{2.0pt}
  \renewcommand{\arraystretch}{1.05}
  \begin{tabularx}{\textwidth}{@{}
    >{\centering\arraybackslash}p{0.12\textwidth}
    >{\hsize=1.15\hsize\raggedright\arraybackslash}X
    >{\hsize=0.85\hsize\raggedright\arraybackslash}X
    c c c c c @{}}
    \toprule
    \textbf{Benchmark} & \textbf{Method} & \textbf{Model} & \multicolumn{4}{c}{\textbf{Subsets}} & \textbf{Overall} \\
    \cmidrule(lr){4-7} \cmidrule(l){8-8}
    & & & \textbf{Chat} & \textbf{Chat-Hard} & \textbf{Safety} & \textbf{Reason} & \\
    \midrule
    \multirow{8}{0.12\textwidth}{\centering\arraybackslash\textit{RewardBench}} & TIR-Judge~\citep{xu2025TIRJudge}$^{\dagger}$ & TIR-Judge-8B & 94.7 & 77.4 & 88.8 & 95.7 & 89.1 \\
    & Agentic Reward Model~\citep{peng2025AgenticRM} & ArmoRM-Llama3-8B$^{\ast}$ & 90.8 & 70.0 & 86.5 & 93.2 & 87.6 \\
    & Multi-Agent Verification~\citep{lifshitz2025multi} & Qwen3-8B & 78.5 & 80.2 & 86.1 & 86.7 & 82.9 \\
    & VerifiAgent~\citep{han2025VerifiAgent} & Qwen3-8B & 69.8 & 65.8 & 78.6 & 61.3 & 68.9 \\
    & Themis~\citep{li2024toolaugmented} & Themis-7B & 82.7 & 42.3 & 45.7 & 37.1 & 52.0 \\
    \cmidrule(l){2-8}
    & RRM~\citep{guo2025rewardreasoning}$^{\dagger}$ & RRM-7B & 87.7 & 70.4 & 80.7 & 90.0 & 82.2 \\
    & RM-R1~\citep{chen2025rmr1}$^{\dagger}$ & RM-R1-Distill-7B & 88.9 & 66.2 & 78.4 & 87.0 & 80.1 \\
    & Think-RM~\citep{hong2025thinkrm}$^{\dagger}$ & Think-RM-7B & 94.4 & 77.9 & 85.2 & 86.4 & 86.0 \\
    \midrule
    & & & \textbf{Chat} & \textbf{Math} & \textbf{Code} & \textbf{Safety} & \\
    \midrule
    \multirow{8}{0.12\textwidth}{\centering\arraybackslash\textit{RM-Bench}} & TIR-Judge~\citep{xu2025TIRJudge}$^{\dagger}$ & TIR-Judge-8B & 80.1 & 91.9 & 69.0 & 93.9 & 83.7 \\
    & Agentic Reward Model~\citep{peng2025AgenticRM} & ArmoRM-Llama3-8B$^{\ast}$ & 60.0 & 58.3 & 54.5 & 86.8 & 64.9 \\
    & Multi-Agent Verification~\citep{lifshitz2025multi} & Qwen3-8B & 62.6 & 67.8 & 52.4 & 84.9 & 66.9 \\
    & VerifiAgent~\citep{han2025VerifiAgent} & Qwen3-8B & 41.3 & 80.0 & 45.9 & 74.5 & 60.4 \\
    & Themis~\citep{li2024toolaugmented} & Themis-7B & 48.4 & 49.9 & 53.9 & 48.4 & 50.2 \\
    \cmidrule(l){2-8}
    & RRM~\citep{guo2025rewardreasoning}$^{\dagger}$ & RRM-7B & 58.4 & 81.8 & 56.7 & 84.9 & 70.4 \\
    & RM-R1~\citep{chen2025rmr1}$^{\dagger}$ & RM-R1-Distill-7B & 64.0 & 83.9 & 56.2 & 85.3 & 72.4 \\
    & Think-RM~\citep{hong2025thinkrm}$^{\dagger}$ & Think-RM-7B & 69.3 & 76.0 & 56.5 & 93.7 & 73.9 \\
    \midrule
    & & & \textbf{Math} & \textbf{Code} & \textbf{Knowledge} & \textbf{Reason} & \\
    \midrule
    \multirow{8}{0.12\textwidth}{\centering\arraybackslash\textit{JudgeBench}} & TIR-Judge~\citep{xu2025TIRJudge}$^{\dagger}$ & TIR-Judge-8B & 71.8 & 70.7 & 60.8 & 71.7 & 66.7 \\
    & Agentic Reward Model~\citep{peng2025AgenticRM} & ArmoRM-Llama3-8B$^{\ast}$ & 63.4 & 49.2 & 47.0 & 56.7 & 51.9 \\
    & Multi-Agent Verification~\citep{lifshitz2025multi} & Qwen3-8B & 54.3 & 42.5 & 45.7 & 39.0 & 44.8 \\
    & VerifiAgent~\citep{han2025VerifiAgent} & Qwen3-8B & 50.0 & 39.6 & 41.6 & 53.9 & 45.5 \\
    & Themis~\citep{li2024toolaugmented} & Themis-7B & 42.9 & 48.0 & 54.8 & 51.6 & 51.1 \\
    \cmidrule(l){2-8}
    & RRM~\citep{guo2025rewardreasoning}$^{\dagger}$ & RRM-7B & 83.2 & 61.9 & 64.3 & 64.2 & 67.0 \\
    & RM-R1~\citep{chen2025rmr1}$^{\dagger}$ & RM-R1-Distill-7B & 82.1 & 71.4 & 64.9 & 62.2 & 67.7 \\
    & Think-RM~\citep{hong2025thinkrm}$^{\dagger}$ & Think-RM-7B & 67.9 & 42.9 & 67.5 & 67.3 & 64.6 \\
    \bottomrule
  \end{tabularx}
  \par\noindent
  \parbox[t]{\textwidth}{\vspace{0.25\baselineskip}\scriptsize
    \textit{Note:} $^{\ast}$Agentic Reward Model uses ArmoRM-Llama3-8B as the base reward model, with Llama3-8B Instruct and Qwen2.5-Coder-7B used inside the verification modules.
    $^{\dagger}$Results are extracted from original-paper comparison tables.
    RewardBench and RM-Bench entries are converted to percentages for consistency with JudgeBench.}
\end{table*}

Table~\ref{tab:evaluation-rm-results} presents side-by-side reward-model scores on RewardBench~\citep{lambert2024rewardbench}, RM-Bench~\citep{liu2024rmbench}, and JudgeBench~\citep{tan2025judgebench} for the agentic methods above~\citep{xu2025TIRJudge,peng2025AgenticRM,lifshitz2025multi,han2025VerifiAgent,li2024toolaugmented} and several non-agentic reasoning reward models~\citep{guo2025rewardreasoning,chen2025rmr1,hong2025thinkrm} included as baselines.
We retain published scores where available and reproduce missing benchmark entries from publicly released source code, since most original papers do not evaluate all three benchmarks.

Table~\ref{tab:evaluation-rm-results} shows no uniform agentic advantage.
On RM-Bench, only TIR-Judge clearly exceeds the non-agentic reasoning reward models; the remaining agentic systems fall below them.
On JudgeBench, strong non-agentic models match or slightly surpass the best agentic entry, while most agentic judges perform less well, particularly outside their strongest categories.
These patterns are consistent with the effectiveness of preference- and reasoning-oriented reward training on fixed response pairs.
They also limit the conclusions that can be drawn about active evidence-seeking, which these comparisons only partly capture.

\subsection{NLG Evaluation}
\label{sec:evaluation-nlg}

NLG evaluators assess generated outputs together with task inputs and any source documents, dialogue context, or references.
Validation measures agreement with human quality, factuality, error, or facet annotations on those outputs rather than downstream task gain.
BLEU, ROUGE, and BERTScore remain common baselines, but only partially align with these judgments in summarization~\citep{fabbri2021summeval}, dialogue~\citep{gopalakrishnan2019topicalchat,mehri2020usr}, and story generation~\citep{mostafazadeh2016rocstories,fan2018hierarchical}.

\begin{table*}[t]
    \centering
    \caption{NLG evaluation benchmarks.}
    \label{tab:evaluation-nlg-benchmarks}
    \scriptsize
    \setlength{\tabcolsep}{2.0pt}
    \renewcommand{\arraystretch}{1.05}
    \begin{tabular*}{\textwidth}{@{\extracolsep{\fill}} >{\raggedright\arraybackslash}p{0.23\textwidth} >{\raggedright\arraybackslash}p{0.11\textwidth} >{\raggedright\arraybackslash}p{0.12\textwidth} >{\raggedright\arraybackslash}p{0.20\textwidth} >{\raggedright\arraybackslash}p{0.16\textwidth} >{\raggedright\arraybackslash}p{0.08\textwidth} C{0.05\textwidth} @{}}
      \toprule
      \textbf{Benchmark / dataset} & \textbf{Evaluand form} & \textbf{Evaluand} & \textbf{Domain} & \textbf{Human signal} & \textbf{Scale} & \textbf{URL} \\
      \midrule
      SummEval~\citep{fabbri2021summeval} & point-wise & summary & summarization & quality rating & 1,600 & \githubartifact{https://github.com/Yale-LILY/SummEval}\,\hfdataset{https://hf.co/datasets/mteb/summeval} \\
      TopicalChat-USR~\citep{gopalakrishnan2019topicalchat,mehri2020usr} & point-wise & response & dialogue & quality rating & 360 & \githubartifact{https://github.com/Shikib/usr} \\
      QAGS~\citep{wang2020qags} & point-wise & summary & summarization & factuality label & 474 & \githubartifact{https://github.com/W4ngatang/qags} \\
      ROCStories~\citep{mostafazadeh2016rocstories,Li2024MATEval} & point-wise & story & story generation & error score & 200 & \hfdataset{https://hf.co/datasets/mintujupally/ROCStories} \\
      WritingPrompts~\citep{fan2018hierarchical,Li2024MATEval} & point-wise & story & story generation & error score & 200 & \hfdataset{https://hf.co/datasets/euclaise/writingprompts} \\
      StorySparkQA~\citep{chen-etal-2024-storysparkqa,chen2025MultiAgentasJudge} & point-wise & QA pair & story QA & expert rating & 70 & \githubartifact{https://github.com/neuhai/StorySparkQA}\,\hfdataset{https://hf.co/datasets/NEU-HAI/StorySparkQA} \\
      MSLR-COCHRANE~\citep{wang-etal-2023-automated,chen2025MultiAgentasJudge} & point-wise & summary & biomedical summarization & facet rating & 102 & \githubartifact{https://github.com/allenai/mslr-annotated-dataset}\,\hfdataset{https://hf.co/datasets/allenai/mslr2022} \\
      \bottomrule
    \end{tabular*}
  \end{table*}

Table~\ref{tab:evaluation-nlg-benchmarks} catalogs datasets by evaluand, domain, and human signal.
Although no single suite spans these settings, they share the goal of recovering human judgments of free-form language quality~\citep{liu2023G-Eval,chan2024ChatEval}.

NLG quality is multidimensional and often under-specified by a single rubric or reference.
Agentic evaluators make that ambiguity inspectable: criteria are separated across agents, borderline cases are debated, rationales are compared, and preliminary scores are revised before aggregation or arbitration.
ChatEval~\citep{chan2024ChatEval} is a representative multi-agent debate instance;
MAJ-EVAL~\citep{chen2025MultiAgentasJudge} and CollabEval~\citep{gupta2026CollabEval} likewise evaluate free-form text against human quality annotations.

Specialized directions within NLG evaluation focus on particular quality dimensions or generation tasks.
Summary factuality systems, including MADISSE~\citep{koupaee2025Faithful} and NarrativeFactScore~\citep{jeong-etal-2025-agent}, assess whether generated summaries are supported by the source, focusing on factual consistency rather than overall summary quality.
Machine-translation evaluators such as HiMATE~\citep{zhang2025HiMATE} and M-MAD~\citep{feng2025MMAD} assess translation quality through MQM and MT-metric targets, with translation errors as the central evaluation signal.

TALE~\citep{TALE} and Multi-Agent LLM Judge~\citep{cao2025MultiAgentLLMJudge} move toward answer correctness, reference consistency, or retrieval support, so QA task datasets used only as answer sources should not be treated as NLG evaluator benchmarks.

Judge-reliability and meta-evaluation work instead examines judgments or judging procedures.
JudgeBench~\citep{tan2025judgebench}, LLMBar, MLLM-Judge, and JudgeAnything test judgment correctness using response pairs, preference labels, or multimodal cases.
They belong closer to the reward-model and judge-quality axis, including when used to study debate, aggregation, or stopping~\citep{braun2025MetaJudges,hu2025AdaptiveStability}.

Reward modeling offers the densest shared comparisons, while NLG reporting remains dataset- and lineage-local.
Both primarily validate preference or annotation agreement; dedicated agentic benchmarks are needed to establish when evidence-seeking improves verdict reliability.

\section{Application}\label{sec:apps}
\begin{figure}[t]
\centering
\includegraphics[width=1\linewidth]{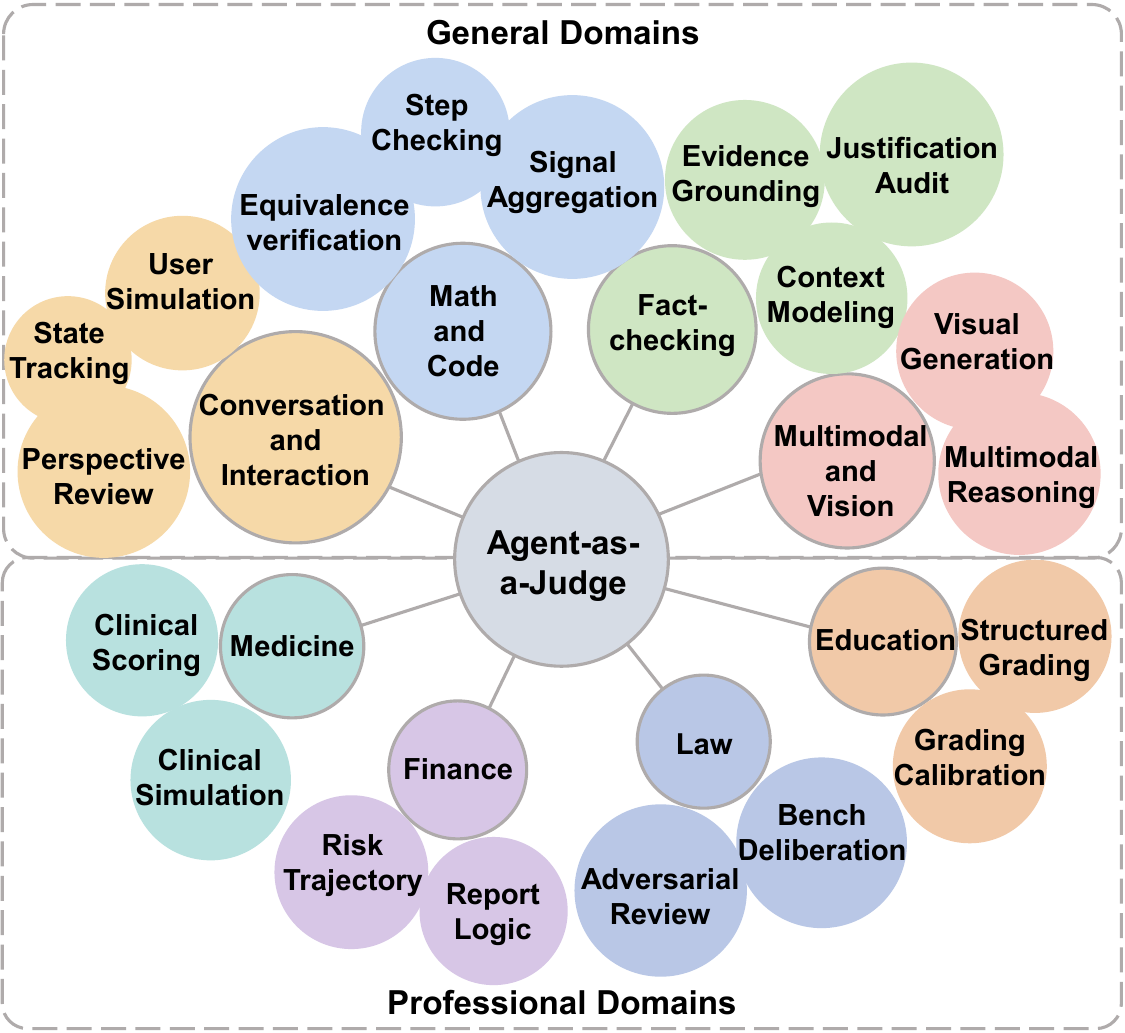}
\caption{Application map of \textit{Agent-as-a-Judge}. General domains mainly stress verifiability, evidence gathering, interaction, and multimodal inspection, while professional domains add stronger requirements on expertise, traceability, safety, and consistency.}
\vspace{-1em}
\label{fig:app}
\end{figure}

Applications extend judgment beyond final text to reasoning, evidence, visual details, interaction histories, and professional constraints.
Figure~\ref{fig:app} maps general and professional domains; Table~\ref{tab:app-gaps} summarizes their judgment targets, fixed-context evaluation gaps, and corresponding agentic capabilities.

\definecolor{appGeneralBg}{HTML}{F5F8FC}
\definecolor{appProfessionalBg}{HTML}{F8FAF4}
\definecolor{appMath}{HTML}{C4D7F2}
\definecolor{appFact}{HTML}{CFE6C3}
\definecolor{appConversation}{HTML}{F6D9A8}
\definecolor{appVision}{HTML}{F4C8C4}
\definecolor{appMedicine}{HTML}{B8E1DF}
\definecolor{appFinance}{HTML}{D7C6E6}
\definecolor{appLaw}{HTML}{B9C7E6}
\definecolor{appEducation}{HTML}{F1C9A8}

\begin{table*}[t]
  \centering
  \caption{Application-specific requirements for \textit{Agent-as-a-Judge}. Each domain is summarized by its judgment target, the limitation of fixed-context judging, and the agentic capability used to address it.}
  \label{tab:app-gaps}
  \scriptsize
  \setlength{\tabcolsep}{5.2pt}
  \renewcommand{\arraystretch}{1.22}
  \begin{tabular}{@{}
    >{\raggedright\arraybackslash}p{2.55cm}
    >{\raggedright\arraybackslash}p{4.0cm}
    >{\raggedright\arraybackslash}p{5.0cm}
    >{\raggedright\arraybackslash}p{5.0cm}
  @{}}
    \toprule
    \textbf{Domain} &
    \textbf{Judgment target} &
    \textbf{Evaluation gap} &
    \textbf{Agentic capability} \\
    \midrule
    \rowcolor{appGeneralBg}[\tabcolsep][\tabcolsep]
    \multicolumn{4}{@{}c@{}}{\textbf{General domains}} \\
    \addlinespace[1pt]
    \textcolor{appMath}{\rule{1.8ex}{1.8ex}}\ Math and code &
    Reasoning steps and executable outputs &
    Correctness may depend on execution, equivalence, or hidden step errors. &
    Execution, solver checks, and multi-verifier agreement \\
    \addlinespace[1pt]
    \textcolor{appFact}{\rule{1.8ex}{1.8ex}}\ Fact-checking &
    Claims, verdicts, and justifications &
    A label is weak without retrieved support and justification checks. &
    Retrieval, evidence grounding, and justification verification \\
    \addlinespace[1pt]
    \textcolor{appConversation}{\rule{1.8ex}{1.8ex}}\ Conversation &
    Multi-turn behavior and responses &
    Quality depends on goals, reactions, and state across turns. &
    User simulation, memory, and multi-perspective review \\
    \addlinespace[1pt]
    \textcolor{appVision}{\rule{1.8ex}{1.8ex}}\ Multimodal and vision &
    Images and multimodal outputs &
    Evidence is often localized in regions, pages, or cross-modal details. &
    Visual inspection, targeted questioning, and multimodal tools \\
    \midrule
    \rowcolor{appProfessionalBg}[\tabcolsep][\tabcolsep]
    \multicolumn{4}{@{}c@{}}{\textbf{Professional domains}} \\
    \addlinespace[1pt]
    \textcolor{appMedicine}{\rule{1.8ex}{1.8ex}}\ Medicine &
    Clinical reports and medical dialogue &
    Errors are safety-sensitive and require expertise and uncertainty handling. &
    Role-specialized evaluation, tool-assisted scoring, and simulation \\
    \addlinespace[1pt]
    \textcolor{appLaw}{\rule{1.8ex}{1.8ex}}\ Law &
    Legal arguments and verdicts &
    Plausible verdicts may miss opposing arguments or disagreement. &
    Adversarial debate, deliberation, and consensus control \\
    \addlinespace[1pt]
    \textcolor{appFinance}{\rule{1.8ex}{1.8ex}}\ Finance &
    Reports, agents, and risk trajectories &
    Judgment must account for report logic, time, and trajectory failures. &
    Planning, logic extraction, and trajectory audit \\
    \addlinespace[1pt]
    \textcolor{appEducation}{\rule{1.8ex}{1.8ex}}\ Education &
    Answers, rubrics, and rationales &
    Scores need consistent rubrics and inspectable grading evidence. &
    Rubric planning, evidence recognition, cross-review, and refinement \\
    \bottomrule
  \end{tabular}
\end{table*}

\subsection{General Domains}\label{sec:apps:general}
General domains emphasize correctness checking, evidence retrieval, interaction simulation, and multimodal inspection.

\paragraph{Math and Code}
1) Step checking inspects intermediate reasoning rather than relying on the final answer alone.
HERMES~\citep{ospanov2025HERMES} anchors mathematical reasoning to intermediate formal proof steps, while VerifiAgent~\citep{han2025VerifiAgent} separates high-level reasoning assessment from tool-based correctness verification.
2) Equivalence verification checks answers with different surface forms.
CompassVerifier~\citep{liu-etal-2025-compassverifier} and xVerify~\citep{chen2025xverify} target this setting by checking mathematical and logical outputs beyond literal string match.
3) Signal aggregation combines multiple checks.
Multi-Agent Verification~\citep{lifshitz2025multi} distributes assessment across aspect-specific judges, Agentic Reward Modeling~\citep{peng2025AgenticRM} combines preference supervision with verifiable correctness signals, and Popper~\citep{huang2025popper} treats judgment as controlled falsification through statistical tests.

\paragraph{Fact-Checking}
1) Evidence grounding retrieves external support before assigning a verdict.
UrduFactCheck~\citep{ahmad-etal-2025-urdufactcheck} uses multilingual retrieval and evidence boosting to improve robustness in low-resource fact checking.
2) Justification audit evaluates both the verdict and the explanation that supports it.
Adaptive audit and dialectical checking assess verdicts against justifications or supporting and refuting evidence before further investigation or closure~\citep{lin-etal-2025-fact,li2026DARE}.
3) Context modeling tracks long-range factual relations.
NarrativeFactScore~\citep{jeong-etal-2025-agent} builds character-level knowledge representations to detect state and relation errors in long narratives.

\paragraph{Conversation and Interaction}
1) User simulation creates controllable interactions for evaluating behavior across turns.
IntellAgent~\citep{levi2025intellagent} uses interactive user simulations to synthesize conversational benchmarks, while the framework of \citet{kazi2024large} introduces controllable user goals and automatic measures.
2) State tracking evaluates whether the system handles affective or social dynamics over time.
ESC-Judge~\citep{madani-srihari-2025-esc} evaluates emotional-support behavior through standardized counseling skills, Sentient Agent~\citep{zhang2025sentient} tracks emotion trajectories, and PSYCHE~\citep{lee2025psyche} builds psychiatric patient profiles for ethical assessment validation.
3) Perspective review combines reviewer positions on subjective and objective quality.
Role-player summarization evaluation~\citep{wu2023large} frames evaluation as multi-perspective role play with diverse reviewer personas.

\paragraph{Multimodal and Vision}
1) Visual generation evaluation checks prompt adherence, controls, and user-defined aspects.
Tool orchestration and multi-round inspection assess control adherence, subject consistency, user-tailored aspects, and spatial constraints~\citep{wang-etal-2025-unified,zhang2025EvaluationAgent,tan2026SpatialCritic}.
2) Multimodal verification grounds claims, reasoning steps, and trajectory events in observable evidence.
Visual questions, targeted image operations, and temporal inspection support image-text fact checking, reasoning rewards, and progress estimation from offline robot trajectories~\citep{beigi2024lrqfact,TIM-PRM,ding2025ARMThinker,hu2026ARS}.

\subsection{Professional Domains} \label{sec:apps:professional}
Professional domains add requirements for expertise, accountability, safety, and traceability to these evaluation needs.

\paragraph{Medicine}
1) Clinical scoring evaluates medical responses or reports through specialized criteria.
MAJ-Eval~\citep{chen2025MultiAgentasJudge} constructs multiple evaluator personas to debate and cross-verify responses, while GEMA-Score~\citep{zhenxuan2025GEMA-Score} uses agent collaboration to compute granular, tool-assisted scores covering disease severity and uncertainty.
2) Clinical simulation evaluates medical behavior through interactive clinical roles.
Chat-Coach~\citep{huang2024benchmarking} pairs patient and coach agents to critique trainee-doctor dialogues, while AI Hospital~\citep{fan2025ai} evaluates LLM doctors in multi-agent clinical simulations, though final scoring may still rely on conventional metrics.

\paragraph{Law}
1) Adversarial review exposes the judge to conflicting legal arguments.
AgentsCourt~\citep{he2024agentscourt} simulates prosecutors, defense attorneys, and judges so that the final judgment is conditioned on opposing positions.
2) Bench deliberation models how legal decision makers negotiate agreement or preserve dissent.
SAMVAD~\citep{devadiga2025samvad} and AgentsBench~\citep{jiang2025agentsbench} simulate judicial deliberation, making concurring and dissenting opinions part of the evaluation process.

\paragraph{Finance}
1) Report logic checks whether a long-form financial analysis follows a coherent research structure.
FinResearchBench~\citep{sun2025finresearchbench} extracts logic trees from analyst reports as intermediate structures for assessment, while FinDeepResearch~\citep{zhu2025findeepresearch} synthesizes hierarchical rubrics for deep financial research evaluation.
2) Risk trajectory audits how financial agents behave over time or across teams.
SAEA~\citep{chen2025standardbenchmarksfail} audits agent trajectories to detect hallucination and temporal misalignment, while From Tasks to Teams~\citep{chen2025tasks} traces multi-agent failures such as cross-agent divergence and error propagation.

\paragraph{Education}
1) Structured grading makes scoring depend on explicit criteria, evidence recognition, and cross-review.
Grade-Like-a-Human~\citep{xie2024grade} and AutoSCORE~\citep{wang2025autoscore} decompose grading into rubric construction, evidence recognition, and score review.
2) Grading calibration diagnoses disagreement and refines the standards used for later grading.
MAJ-Eval~\citep{chen2025MultiAgentasJudge} uses multi-persona debate to approximate multidimensional human evaluation, while GradeOpt~\citep{chu2024llm} introduces agents that diagnose discrepancies and iteratively refine grading guidelines.

\section{Discussion}\label{sec:discussion}
The mechanisms that expand what agentic judges can inspect also introduce deployment and governance challenges.
We examine these challenges and directions for making evaluation selective, reliable, and controllable.

\subsection{Challenges}

\paragraph{Efficiency}
Additional evaluation actions increase cost and latency.
Training tool use, planning, or adaptive routing may require trajectory data or reinforcement learning rather than simple supervised labels.
At inference time, reasoning, tool calls, memory access, and multi-agent coordination incur costs that compound when later checks depend on earlier observations.
The resulting latency limits online evaluation, moderation, and reinforcement-learning feedback loops.
Efficiency therefore depends on balancing the expected value of further checking against its computational and interaction costs.

\paragraph{Process Reliability}
Errors can arise from intermediate actions as well as judge-model bias: planning may target the wrong claim, retrieval may supply weak evidence, tool results may be misinterpreted, and deliberation may converge on an incorrect consensus.
Memory can preserve these mistakes across steps or cases.
A verdict may consequently appear well supported despite failures in its evidence trace.
Reliability assessment must therefore examine plans, observations, memory updates, and resolution rules alongside final verdicts.

\paragraph{Safety and Control}
Access to search engines, executors, databases, and interactive environments exposes judges to prompt injection, tool misuse, and unintended side effects.
Multi-agent protocols can propagate unsafe rationales or adversarial instructions through shared outputs.
When verdicts become reward signals, downstream optimization may reinforce systematic judging errors.
Deployment therefore requires bounded tool permissions and external actions, explicit stopping rules, and escalation paths when automated evaluation cannot safely resolve a case.

\paragraph{Privacy and Governance}
Persistent evaluation state may contain user information, evidence, prior verdicts, or personalized criteria.
Reuse can leak sensitive records, transfer user-specific standards across contexts, or perpetuate outdated judgments, particularly in professional domains involving confidential or identifiable information.
Governance must specify what may be stored, retention periods, access rights, deletion requirements, and permitted reuse across cases.

\subsection{Future Directions}
These challenges motivate advances in evaluation control, memory management, evidence creation, and learned judging policies.

\paragraph{Adaptive Control}
Judges should allocate evaluation effort according to the case: use lightweight assessment when sufficient, decompose difficult cases, and escalate weak evidence to additional tools, agents, or human review.
This requires adapting criteria, granularity, and evidence depth while accounting for both the cost and error risk of further actions.
Generalization thus includes transferring control strategies across cases rather than applying a fixed rubric and evaluation budget.

\paragraph{Memory Management}
Memory management should become an explicit part of the judging policy, with decisions about storage, summarization, retrieval, revision, and forgetting.
The aim is to support personalization, consistency, and cross-case learning while keeping evidence current and reuse appropriate to the target case.
In professional domains, these decisions must also respect confidentiality and distinguish reusable standards from case-specific information.

\paragraph{Autonomous Probing}
Current probing through tool calls, visual questions, simulations, and environment interaction is often local, predefined, or tied to narrow interfaces.
Future judges could generate targeted tests, counterexamples, alternative prompts, environment states, or stress cases when the initial context is insufficient.
The challenge is to select probes that reduce judgment uncertainty or expose specific failure boundaries, making evidence creation purposeful rather than merely extending the evaluation trace.

\paragraph{Trainable Judgment Policies}
Learning could replace hand-designed prompts, workflows, and routing rules with policies for decomposition, tool selection, evidence sufficiency, deliberation, and stopping.
Reinforcement learning and process-level supervision offer training signals that can account for verdict quality together with trace cost and reliability.
The central difficulty is stability: improvements in planning, tool use, and coordination must preserve controllability and auditability.

Progress should be measured by selective, evidence-grounded, and accountable evaluation: whether the judge seeks appropriate evidence, preserves useful state, and stops or escalates at the right point.

\section{Conclusion}

This survey organizes \textit{Agent-as-a-Judge} around the construction and control of evaluation processes. We established operational criteria for evaluation-directed checks and the use of their results, and connected multi-agent collaboration, tool integration, memory, and planning to the findings on which a verdict rests.
We further mapped validation settings, benchmark coverage, and application requirements across general and professional domains.
While promising, this evolution presents challenges in computational cost, latency, safety, and privacy. Future progress should prioritize personalization, generalization, and optimization, ultimately realizing truly autonomous evaluators that continuously adapt to the evolving AI landscape.

\bibliographystyle{unsrtnat}
\bibliography{references}

\clearpage
\begin{center}
  {\LARGE\bfseries Supplementary Material\par}
\end{center}
\phantomsection
\label{app:supp}

This appendix collects material that supports the survey beyond the main text.

\appendices
\section{Paper Inclusion Criteria}
\label{app:inclusion}

Section~\ref{sec:background} fixes the conceptual boundary of \textit{Agent-as-a-Judge}; this appendix makes the corresponding screening standard explicit.
Screening uses the reported checking procedure, its intermediate results, and their role in verdict formation.
The mechanism-specific descriptions below follow the organization of Section~\ref{sec:methods}.

\paragraph{Working definition}
An \emph{agentic judge} constructs a verdict about an evaluand through explicit checking operations and the use of their results.
Qualifying operations obtain evidence, perform targeted verification, or reassess a judgment using another check's findings or rationale, beyond replicating an overall score.
They may follow a prescribed workflow or adapt to intermediate observations.

\paragraph{Minimal inclusion rule}
A paper is included when all three conditions hold:
\textbf{(1) Evaluation purpose.} Its main contribution is a judge, verifier, reward evaluator, or evaluation procedure that forms a verdict about a specified evaluand.
\textbf{(2) Substantive checking.} At least one explicit operation obtains evidence, tests a claim or constraint, or reassesses a judgment using another check's findings or rationale. Replicated overall scoring, including independent persona variants, does not satisfy this condition on its own.
\textbf{(3) Result use.} The reported procedure specifies how the check result is consumed by subsequent assessment or verdict formation. Qualifying results include retrieved passages, execution outcomes, claim-specific verification decisions, extracted discrepancies, and critiques used for reassessment.
Counting independent overall scores or labels alone does not satisfy conditions~(2)--(3).

\paragraph{Screening record}
The unit of assessment is a reported system configuration or protocol.
For each qualifying process, the screening record identifies the checked object, the checking operation, its output, the downstream consumer, and the source passage or algorithm establishing that dependency.
For delegated verification, the checked proposition, constraint, or comparison must be identifiable; changing a persona name or requesting another generic dimension score is insufficient.
Inclusion is based on the reported use of results, without requiring that the verdict change on every input.
Configurations with insufficient procedural detail are marked unresolved rather than inferred from an agent or tool label.

\begin{table*}[t]
  \centering
  \caption{Operational boundary examples. These are protocol patterns; a paper may report more than one configuration.}
  \label{tab:inclusion-boundary}
  \small
  \setlength{\tabcolsep}{5pt}
  \renewcommand{\arraystretch}{1.15}
  \begin{tabularx}{\textwidth}{@{}p{0.28\textwidth}p{0.10\textwidth}X@{}}
    \toprule
    Protocol pattern & In scope? & Decisive procedural evidence \\
    \midrule
    One-shot scoring or CoT alone & No & No separate checking operation and result-use dependency. \\
    Independent overall scores followed by voting or averaging & No & Replicated verdicts are counted without further verification or reassessment. \\
    Different personas independently score the same object & No, alone & Role descriptions change, but no distinct verification task or use of peer findings is established. \\
    Fixed-round critique and reassessment & Yes & A later assessment consumes another evaluator's findings or rationale before aggregation. \\
    Independent targeted verifiers followed by aggregation & Yes & Each verifier checks an identifiable proposition or constraint; the resolver uses those verification results. \\
    Fixed retrieval or execution before judgment & Yes & Retrieved evidence or test outcomes are used to assess the evaluand. \\
    Plan generation without executed checks & No, alone & The plan is produced but no corresponding checking results are used. \\
    Findings trigger a new check, a revised query, or stopping & Yes & Both a qualifying check and observation-driven control are explicit. \\
    \bottomrule
  \end{tabularx}
\end{table*}

\paragraph{Control mode}
For an included configuration, control is characterized separately as prescribed execution, input-conditioned planning, or observation-driven revision.
A fixed rule can implement observation-driven control when a returned finding triggers a branch, further check, escalation, or stopping.
Conversely, input-specific plans or changing discussion content do not by themselves establish revision of the checking procedure.

Table~\ref{tab:inclusion-boundary} contrasts common protocols; the following descriptions show how the four mechanisms realize the inclusion conditions.

\paragraph{Planning}
The judge constructs or selects an explicit plan, rubric, or decomposition that organizes subsequent checks, whose results inform the verdict.
The plan may be executed as a prescribed workflow or revised in response to critiques, tests, counterexamples, or other intermediate findings.
A plan expressed only as reasoning text, without corresponding checking operations, does not qualify on its own.

\paragraph{Tool integration}
1) \emph{Tool use.} The judge invokes external tools to gather evidence---search or retrieval, compilers or executors, provers, browsers or GUIs, databases---and tool results can change the verdict.
2) \emph{Interaction with external state.} The judge probes or acts on files, a GUI, a simulator, the web, or executable artifacts to obtain observations used in evaluation.
Tool calls may follow a prescribed sequence or be selected from intermediate findings; the external state itself need not change during evaluation.

\paragraph{Multi-agent collaboration}
The judge delegates identifiable verification tasks or coordinates the use of peer findings and rationales in subsequent assessment.
The decisive evidence is the checking responsibility or information dependency, rather than role diversity or the final aggregation operator.
Prescribed critique exchanges can satisfy this condition even when the speaking order and number of rounds are fixed; parallel verifiers can satisfy it when they test distinct propositions or constraints rather than replicate overall scoring.
For example, peer-informed reassessment appears in debate and collaboration protocols~\citep{chan2024ChatEval,gupta2026CollabEval}, while structured clinical comparison~\citep{zhenxuan2025GEMA-Score} and aspect verification~\citep{lifshitz2025multi} organize distinct checking tasks.

\paragraph{Memory}
The system maintains an explicit write--manage--read loop over a memory store that affects later actions or the verdict, rather than relying only on a longer context window, a fixed rubric in the prompt, or a chain-of-thought scratchpad.
The reported procedure must specify how stored findings or artifacts are accessed and used in evaluation.

\paragraph{Non-examples}
The following patterns do not qualify on their own:
1) multi-step reasoning in a single forward pass;
2) a generated plan or rubric that is not used to organize subsequent checking operations;
3) replicated overall scoring, including independent persona variants, or self-consistency sampling followed only by voting;
4) appending full history into context without a memory loop;
5) intermediate artifacts that are displayed but not used in subsequent assessment or verdict formation.

\paragraph{Common mis-inclusion cases}
1) \emph{Standalone formal verification.} A prover alone does not constitute an LLM-based judge; an evaluator that invokes a prover and uses its results is assessed by the same checking and result-use conditions.
2) \emph{RL training methods or infrastructure.} Work whose main contribution is reinforcement-learning training or tooling rather than the judge architecture.
3) \emph{Judge as a sub-component.} Systems where judging is ancillary to a larger pipeline.

\section{Construction of MAD Subagent Figures}
\label{app:mad-figures}

\newcommand{\madnames}[1]{\begin{tabular}[t]{@{}p{0.34\textwidth}@{}}#1\end{tabular}}
\newcommand{\madtags}[1]{\begin{tabular}[t]{@{}l@{}}#1\end{tabular}}
\newcommand{\madtag}[1]{\texttt{#1}}
\newcommand{\madsource}[2]{\textit{#1}~\citep{#2}}

\paragraph{Scope and record construction}
This appendix documents the manual coding behind Figures~\ref{fig:mad-role-wordclouds} and~\ref{fig:mad-role-tech-sankey}.
The main text states how to read those figures; Table~\ref{tab:mad-figure-sources} lists every plotted record for spot-checking inclusion and axis assignments.
The figures draw on the MAD judge papers analyzed in Section~\ref{sec:methods}: 24 papers and 102 role-level records.
Each record names one active judging component in a paper's protocol; one paper may contribute several records when it names several such positions.
During manual review, paper-specific evidence sentences guide whether a named entity is retained and how it is coded on both axes.
The functional and technical axes follow the definitions used with those figures in the multi-agent section.
Table~\ref{tab:mad-figure-sources} reports them in abbreviated form in the Func.\ and Tech.\ columns; the first table caption defines the abbreviations.
We retain only components that produce, check, discuss, aggregate, route, or authorize a judgment within the evaluation run.

\begin{table*}[t]
  \centering
  \tiny
  \setlength{\tabcolsep}{2pt}
  \caption{Manually identified MAD subagent/component records used for figure construction, part 1 of 3.
  Func. abbreviates functional role: \madtag{Adv.} = adversarial, \madtag{Persp.} = perspective, \madtag{Analyt.} = analytical, and \madtag{V-Ctrl.} = verdict-control.
  Tech. abbreviates implementation status: \madtag{Prompt} = persona prompts, \madtag{Workflow} = structured workflows, \madtag{Ctrl.} = deterministic controllers, and \madtag{Tool/env.} = tool-integrated components.}
  \label{tab:mad-figure-sources}
  \resizebox{\textwidth}{!}{
  \begin{tabular}{@{}p{0.20\textwidth}p{0.36\textwidth}p{0.12\textwidth}p{0.22\textwidth}@{}}
    \toprule
    Source paper & Subagent/component & Func. & Tech. \\
    \midrule
    \madsource{Table-Critic}{yu2025TableCritic} & \madnames{Judge Agent\\Critic Agent\\Refiner Agent\\Curator Agent} & \madtags{\madtag{V-Ctrl.}\\\madtag{Adv.}\\\madtag{Analyt.}\\\madtag{Analyt.}} & \madtags{\madtag{Tool/env.}\\\madtag{Tool/env.}\\\madtag{Workflow}\\\madtag{Workflow}} \\
    \madsource{Meta-Judges}{braun2025MetaJudges} & \madnames{Meta-judge agents\\Expert role\\Critic role\\General Public role\\Mediator or summarization role} & \madtags{\madtag{Analyt.}\\\madtag{Persp.}\\\madtag{Adv.}\\\madtag{Persp.}\\\madtag{V-Ctrl.}} & \madtags{\madtag{Workflow}\\\madtag{Prompt}\\\madtag{Prompt}\\\madtag{Prompt}\\\madtag{Workflow}} \\
    \madsource{Judging with Many Minds}{ma-etal-2025-judging} & \madnames{General Public agent\\Critic agent\\Judge pool agents\\Meta-Judge\\PINE bias-free agent} & \madtags{\madtag{Persp.}\\\madtag{Adv.}\\\madtag{Persp.}\\\madtag{V-Ctrl.}\\\madtag{Analyt.}} & \madtags{\madtag{Prompt}\\\madtag{Tool/env.}\\\madtag{Ctrl.}\\\madtag{Prompt}\\\madtag{Prompt}} \\
    \madsource{Adaptive Stability Detection}{hu2025AdaptiveStability} & \madnames{Judge agents\\Adaptive stability detector\\Majority-vote fallback} & \madtags{\madtag{Persp.}\\\madtag{V-Ctrl.}\\\madtag{V-Ctrl.}} & \madtags{\madtag{Prompt}\\\madtag{Ctrl.}\\\madtag{Ctrl.}} \\
    \madsource{Efficient Safety MAD}{wu2025EfficientSafetyMAD} & \madnames{Critic Agent\\Defender Agent\\Judge Agent} & \madtags{\madtag{Adv.}\\\madtag{Adv.}\\\madtag{V-Ctrl.}} & \madtags{\madtag{Prompt}\\\madtag{Prompt}\\\madtag{Tool/env.}} \\
    \madsource{CollabEval}{gupta2026CollabEval} & \madnames{Evaluator agents\\Discussion participants\\Consensus checker/control logic\\Final judge} & \madtags{\madtag{Analyt.}\\\madtag{Persp.}\\\madtag{V-Ctrl.}\\\madtag{V-Ctrl.}} & \madtags{\madtag{Workflow}\\\madtag{Prompt}\\\madtag{Ctrl.}\\\madtag{Workflow}} \\
    \madsource{AgentV-RL}{AgentV-RL} & \madnames{Forward agent\\Backward agent} & \madtags{\madtag{Analyt.}\\\madtag{V-Ctrl.}} & \madtags{\madtag{Tool/env.}\\\madtag{Prompt}} \\
    \madsource{Multi-Agent LLM Judge}{cao2025MultiAgentLLMJudge} & \madnames{Sample Selection agent\\Evaluation agent\\ReWrite agent} & \madtags{\madtag{Persp.}\\\madtag{V-Ctrl.}\\\madtag{V-Ctrl.}} & \madtags{\madtag{Prompt}\\\madtag{Prompt}\\\madtag{Workflow}} \\
    \bottomrule
  \end{tabular}
  }
\end{table*}

\begin{table*}[t]
  \centering
  \tiny
  \setlength{\tabcolsep}{2pt}
  \caption{Manually identified MAD subagent/component records used for figure construction, part 2 of 3.}
  \resizebox{\textwidth}{!}{
  \begin{tabular}{@{}p{0.20\textwidth}p{0.36\textwidth}p{0.12\textwidth}p{0.22\textwidth}@{}}
    \toprule
    Source paper & Subagent/component & Func. & Tech. \\
    \midrule
    \madsource{ChatEval}{chan2024ChatEval} & \madnames{Debater/referee agents\\General Public\\Critic\\News Author\\Psychologist\\Scientist\\Summarizer} & \madtags{\madtag{Adv.}\\\madtag{Persp.}\\\madtag{Adv.}\\\madtag{Persp.}\\\madtag{Persp.}\\\madtag{Persp.}\\\madtag{V-Ctrl.}} & \madtags{\madtag{Prompt}\\\madtag{Prompt}\\\madtag{Prompt}\\\madtag{Prompt}\\\madtag{Prompt}\\\madtag{Prompt}\\\madtag{Workflow}} \\
    \madsource{Multi-Agent-as-Judge}{chen2025MultiAgentasJudge} & \madnames{Stakeholder agents\\In-group moderator\\Out-group/impartial aggregator\\Persona creation LLM} & \madtags{\madtag{Persp.}\\\madtag{Persp.}\\\madtag{V-Ctrl.}\\\madtag{Persp.}} & \madtags{\madtag{Workflow}\\\madtag{Prompt}\\\madtag{Workflow}\\\madtag{Workflow}} \\
    \madsource{SAMVAD}{devadiga2025samvad} & \madnames{Judge Agent\\Prosecution Counsel Agent\\Defense Counsel Agent\\Adjudicator Agents\\Orchestrator} & \madtags{\madtag{V-Ctrl.}\\\madtag{Adv.}\\\madtag{Adv.}\\\madtag{Analyt.}\\\madtag{V-Ctrl.}} & \madtags{\madtag{Tool/env.}\\\madtag{Tool/env.}\\\madtag{Tool/env.}\\\madtag{Prompt}\\\madtag{Ctrl.}} \\
    \madsource{M-MAD}{feng2025MMAD} & \madnames{Accuracy expert\\Fluency expert\\Terminology expert\\Style expert\\Debater A\\Debater B\\Consensus checker\\Final judge} & \madtags{\madtag{Persp.}\\\madtag{Persp.}\\\madtag{Persp.}\\\madtag{Persp.}\\\madtag{Adv.}\\\madtag{Adv.}\\\madtag{V-Ctrl.}\\\madtag{V-Ctrl.}} & \madtags{\madtag{Prompt}\\\madtag{Prompt}\\\madtag{Prompt}\\\madtag{Prompt}\\\madtag{Prompt}\\\madtag{Prompt}\\\madtag{Ctrl.}\\\madtag{Prompt}} \\
    \madsource{AgentsCourt}{he2024agentscourt} & \madnames{Judge agent\\Plaintiff agent\\Defendant agent\\Judge Assistant agent} & \madtags{\madtag{V-Ctrl.}\\\madtag{Adv.}\\\madtag{Adv.}\\\madtag{Analyt.}} & \madtags{\madtag{Tool/env.}\\\madtag{Prompt}\\\madtag{Prompt}\\\madtag{Tool/env.}} \\
    \madsource{AgentsBench}{jiang2025agentsbench} & \madnames{Presiding professional judge\\Professional judge agents\\Lay judge or juror agents} & \madtags{\madtag{V-Ctrl.}\\\madtag{Persp.}\\\madtag{Persp.}} & \madtags{\madtag{Ctrl.}\\\madtag{Ctrl.}\\\madtag{Ctrl.}} \\
    \madsource{MATEval}{Li2024MATEval} & \madnames{Evaluator Agent\\Feedback Agent\\Summarizer Agent} & \madtags{\madtag{Analyt.}\\\madtag{V-Ctrl.}\\\madtag{V-Ctrl.}} & \madtags{\madtag{Prompt}\\\madtag{Prompt}\\\madtag{Workflow}} \\
    \madsource{Multi-agent verification}{lifshitz2025multi} & \madnames{Aspect Verifiers\\Verification aggregator/selector\\Verifier engineer or validation selection procedure} & \madtags{\madtag{Analyt.}\\\madtag{V-Ctrl.}\\\madtag{Analyt.}} & \madtags{\madtag{Prompt}\\\madtag{Ctrl.}\\\madtag{Prompt}} \\
    \bottomrule
  \end{tabular}
  }
\end{table*}

\begin{table*}[t]
  \centering
  \tiny
  \setlength{\tabcolsep}{2pt}
  \caption{Manually identified MAD subagent/component records used for figure construction, part 3 of 3.}
  \resizebox{\textwidth}{!}{
  \begin{tabular}{@{}p{0.20\textwidth}p{0.36\textwidth}p{0.12\textwidth}p{0.22\textwidth}@{}}
    \toprule
    Source paper & Subagent/component & Func. & Tech. \\
    \midrule
    \madsource{FACT-AUDIT}{lin-etal-2025-fact} & \madnames{Appraiser agent\\Inquirer agent\\Quality Inspector agent\\Evaluator agent\\Prober agent} & \madtags{\madtag{V-Ctrl.}\\\madtag{Analyt.}\\\madtag{Analyt.}\\\madtag{Analyt.}\\\madtag{Analyt.}} & \madtags{\madtag{Prompt}\\\madtag{Tool/env.}\\\madtag{Workflow}\\\madtag{Prompt}\\\madtag{Workflow}} \\
    \madsource{MATH-VF}{MATH-VF} & \madnames{Formalizer\\Critic\\SymPy-oriented tool agent\\Z3-solver-oriented tool agent} & \madtags{\madtag{Analyt.}\\\madtag{Adv.}\\\madtag{Analyt.}\\\madtag{Analyt.}} & \madtags{\madtag{Workflow}\\\madtag{Workflow}\\\madtag{Tool/env.}\\\madtag{Tool/env.}} \\
    \madsource{No-human-in-the-loop}{No-human-in-the-loop} & \madnames{CI Pattern Audit Agent\\Recommendation Issue Audit Agent\\Report Generation Agent\\Multiple LLM judges\\Consensus synthesizer/conflict policy} & \madtags{\madtag{Analyt.}\\\madtag{Analyt.}\\\madtag{Analyt.}\\\madtag{V-Ctrl.}\\\madtag{V-Ctrl.}} & \madtags{\madtag{Prompt}\\\madtag{Prompt}\\\madtag{Ctrl.}\\\madtag{Ctrl.}\\\madtag{Ctrl.}} \\
    \madsource{CodeVisionary}{wang2025CodeVisionary} & \madnames{Central-control LLM agent\\Context distillation agent\\Multiple judge agents\\Report generator/summarizer} & \madtags{\madtag{Analyt.}\\\madtag{Analyt.}\\\madtag{Persp.}\\\madtag{V-Ctrl.}} & \madtags{\madtag{Tool/env.}\\\madtag{Tool/env.}\\\madtag{Ctrl.}\\\madtag{Workflow}} \\
    \madsource{Diverse role-players}{wu2023large} & \madnames{General Public Judger\\Critic Judger\\News Author Judger\\Dynamic subjective users\\Aggregator} & \madtags{\madtag{Persp.}\\\madtag{Persp.}\\\madtag{Persp.}\\\madtag{Persp.}\\\madtag{Persp.}} & \madtags{\madtag{Workflow}\\\madtag{Prompt}\\\madtag{Prompt}\\\madtag{Prompt}\\\madtag{Ctrl.}} \\
    \madsource{Grade like a human}{xie2024grade} & \madnames{Rubric generation agent/stage\\Grading agent/stage\\Post-grading review agent/stage\\Sampling component} & \madtags{\madtag{Analyt.}\\\madtag{V-Ctrl.}\\\madtag{V-Ctrl.}\\\madtag{Analyt.}} & \madtags{\madtag{Ctrl.}\\\madtag{Workflow}\\\madtag{Ctrl.}\\\madtag{Ctrl.}} \\
    \madsource{HiMATE}{zhang2025HiMATE} & \madnames{Tier-1 MQM category agents\\Subtype evaluator role\\Self-reflection/correction role\\Collaborative discussion role\\Weighted scoring component} & \madtags{\madtag{Analyt.}\\\madtag{Analyt.}\\\madtag{Analyt.}\\\madtag{Analyt.}\\\madtag{V-Ctrl.}} & \madtags{\madtag{Prompt}\\\madtag{Workflow}\\\madtag{Prompt}\\\madtag{Prompt}\\\madtag{Ctrl.}} \\
    \madsource{GEMA-Score}{zhenxuan2025GEMA-Score} & \madnames{Entity Extraction Agent\\Objective Clinical Accuracy Agent\\Subjective Expressiveness Evaluation Agent\\Score Evaluation Agent} & \madtags{\madtag{Analyt.}\\\madtag{Persp.}\\\madtag{Persp.}\\\madtag{V-Ctrl.}} & \madtags{\madtag{Workflow}\\\madtag{Prompt}\\\madtag{Prompt}\\\madtag{Workflow}} \\
    \midrule
    Total & 102 records & & \\
    \bottomrule
  \end{tabular}
  }
\end{table*}

\paragraph{Role lexicon figure}
Figure~\ref{fig:mad-role-wordclouds} assigns each record one functional role and one manually chosen display term kept close to the source paper's role vocabulary.
Each plotted record contributes one count to its functional group; term size within a panel reflects how often that display term appears among records assigned to that group.
At render time, each display term is tokenized on whitespace, and the generic tokens \texttt{judge}, \texttt{role}, and \texttt{agent} are removed so panel emphasis falls on distinctive role wording rather than shared MAD naming conventions.

\paragraph{Role-to-implementation Sankey}
Figure~\ref{fig:mad-role-tech-sankey} uses the same 102 records but counts functional-role/implementation-status pairs rather than display-term frequencies.
Implementation status is coded from how the component is realized in the protocol, not from its role name alone.
\emph{Persona prompts} mark a fixed persona or evaluation dimension without adaptive procedure selection.
\emph{Structured workflows} mark LLM steps coupled to specific upstream and downstream artifacts.
\emph{Deterministic controllers} mark rule-based logic that routes, stops, or aggregates the run.
\emph{Tool-integrated components} mark positions whose judging action invokes tools or external evidence through a predefined interface.
Table abbreviations for these classes appear in the first table caption.

\section{Construction of the Tool-Interface Count Figure}
\label{app:tool-counts}

\paragraph{Scope and counting unit}
This appendix documents the manual coding behind Figure~\ref{fig:tool-overview}.
The main text states how to read that figure; Table~\ref{tab:tool-count-sources} lists every counted paper--interface assignment for spot-checking inclusion and category labels.

The figure summarizes a manually coded subset of the tool-augmented judges discussed in the survey: 34 papers and 60 counted interface assignments. It is not an exhaustive count of all cited methods.
Each assignment links one paper to one interface category; a paper may receive several assignments when its judging protocol uses several external interfaces, but at most one assignment per category.
The figure denominator is therefore 60 counted interface uses, not 34 papers.
The Share column reports each category count divided by 60; it is not the share of papers using that interface.

During manual review, paper-specific descriptions of the judging protocol guide whether an external interface is retained and which category it receives.
We count only interfaces that the judge can call during judgment production, evidence gathering, checking, or interaction.
We exclude tools used only by the evaluand or candidate generator, offline preprocessing outside the judging loop, and papers without tool-augmented judging interfaces.

\begin{table}[t]
  \centering
  \tiny
  \setlength{\tabcolsep}{2pt}
  \caption{Counted tool-interface assignments for Figure~\ref{fig:tool-overview}.
  Share is each category count divided by the 60 counted interface uses.
  Interface abbreviations match the figure legend.}
  \label{tab:tool-count-sources}
  \resizebox{\linewidth}{!}{
  \begin{tabular}{@{}p{0.07\textwidth}p{0.04\textwidth} p{0.04\textwidth} p{0.23\textwidth}@{}}
    \toprule
    Interface & Count & Share & Papers \\
    \midrule
    Code Exec & 17 & 28.3\% & \citep{peng2025AgenticRM,zhuge2025AgentasaJudge,wang2025CodeVisionary,han2025VerifiAgent,xu2025TIRJudge,xu2025ARJudge,AgentV-RL,cosineverifier,RubricIsAllYouNeed,T1,tool-annotator,auto-eval-judge,TATTOO,Agent0-VL,Agentic-Rubrics,VAGEN,agentic-verifier} \\
    Web Search & 13 & 21.7\% & \citep{lin-etal-2025-fact,ahmad-etal-2025-urdufactcheck,peng2025AgenticRM,zhuge2025AgentasaJudge,han2025VerifiAgent,he2024agentscourt,li2024toolaugmented,OpenReward,TALE,tool-annotator,auto-eval-judge,DREAM,wang2025CodeVisionary} \\
    Local Search & 8 & 13.3\% & \citep{jeong-etal-2025-agent,zhuge2025AgentasaJudge,he2024agentscourt,devadiga2025samvad,Med-TIV,T1,auto-eval-judge,TATTOO} \\
    Vision Model & 7 & 11.7\% & \citep{wang-etal-2025-unified,zhang2025EvaluationAgent,GUI-PRA,wang2025CodeVisionary,MMRL,TIM-PRM,VAGEN} \\
    Solver & 5 & 8.3\% & \citep{han2025VerifiAgent,li2024toolaugmented,MATH-VF,Agent0-VL,auto-eval-judge} \\
    Image Ops & 3 & 5.0\% & \citep{wang-etal-2025-unified,ding2025ARMThinker,wang2025CodeVisionary} \\
    Shell & 3 & 5.0\% & \citep{wang2025CodeVisionary,Agentic-Rubrics,VAGEN} \\
    Computer Use & 2 & 3.3\% & \citep{wang2025CodeVisionary,VAGEN} \\
    Translation & 2 & 3.3\% & \citep{ahmad-etal-2025-urdufactcheck,li2024toolaugmented} \\
    \midrule
    Total & 60 & 100\% & 34 distinct papers \\
    \bottomrule
  \end{tabular}
  }
\end{table}

\paragraph{Interface categories}
Each row in Table~\ref{tab:tool-count-sources} uses one of nine interface categories below.
Longer label text appears in the figure legend; the coding rules state what qualifies for each abbreviated label.
\emph{Code Exec} covers general programmatic execution through an interpreter or sandboxed runtime during judging.
\emph{Web Search} covers open-web retrieval or live search APIs.
\emph{Local Search} covers fixed corpora, structured stores, knowledge graphs, indexed artifacts, or lookup in input tables rather than open-web search.
\emph{Vision Model} covers vision-language, OCR, segmentation, or parsing models invoked as perception tools.
\emph{Solver} covers dedicated symbolic solvers, theorem provers, or calculator interfaces counted separately from general code execution.
\emph{Image Ops} covers deterministic image operations such as crop, zoom, diff, or overlay that produce inspectable visual evidence.
\emph{Shell} covers terminal, bash, or shell-level command execution.
\emph{Computer Use} covers GUI or environment interaction beyond a single static capture, including scripted UI actions or computer-use agents.
\emph{Translation} covers machine-translation interfaces used while judging cross-lingual inputs.

\section{Complete Evaluation--Benchmark Coverage Map}
\label{app:evaluation-benchmark-map}

\begin{figure*}[t]
  \centering
  \includegraphics[width=\textwidth,height=0.88\textheight,keepaspectratio]{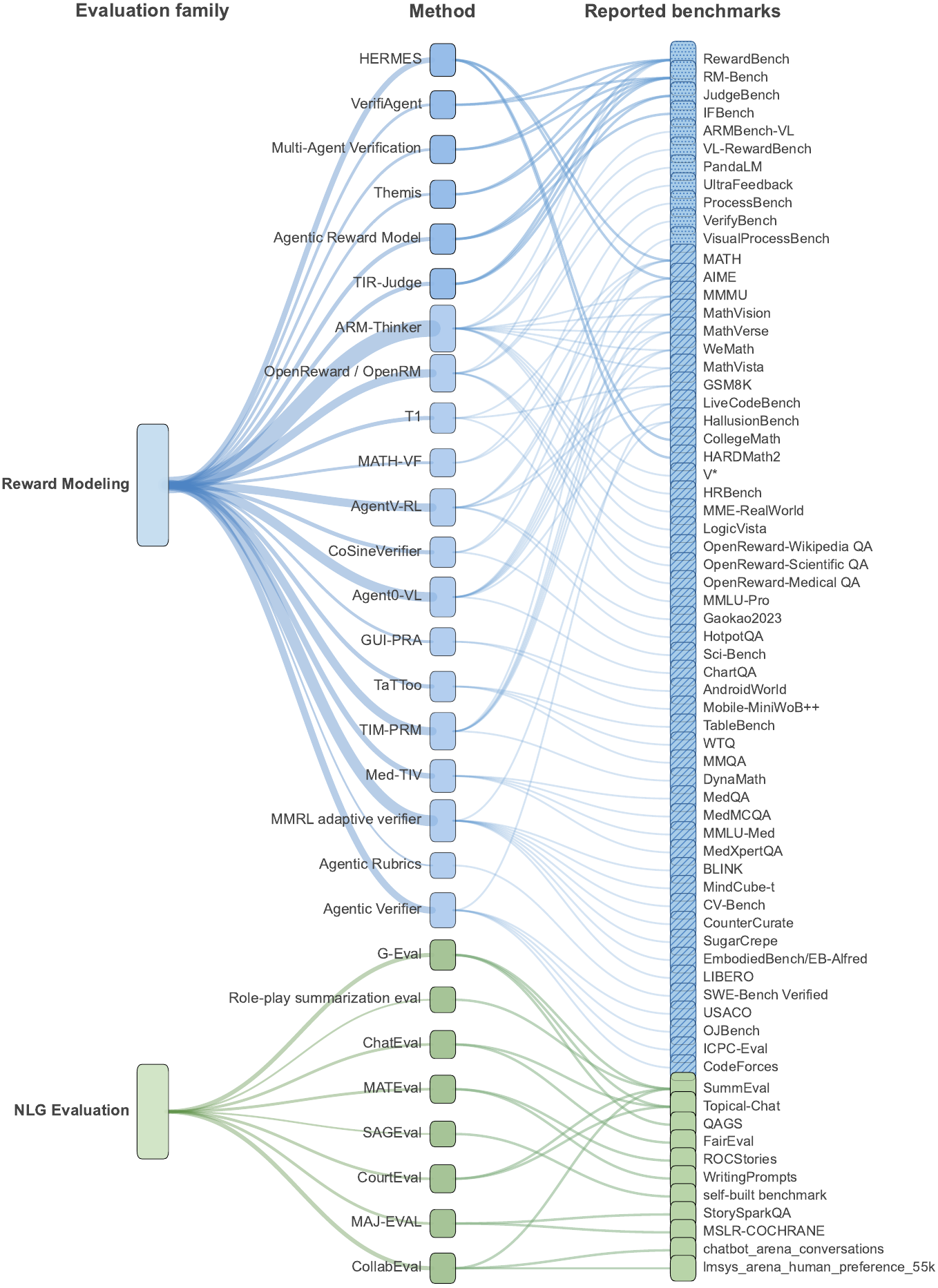}
  \caption{Complete evaluation--benchmark coverage map.
  Unlike Figure~\ref{fig:evaluation-benchmark-map}, this version retains all 100 coded links and singleton benchmark nodes.
  Links indicate that a method reports results on, or is explicitly evaluated on, a reported benchmark.
  Right-node fill and pattern follow Figure~\ref{fig:evaluation-benchmark-map}; see that caption for the artifact-type legend.}
  \label{fig:evaluation-benchmark-map-full}
\end{figure*}

\paragraph{Scope and counting unit}
This appendix documents the manual coding behind Figure~\ref{fig:evaluation-benchmark-map} and Figure~\ref{fig:evaluation-benchmark-map-full}.
The main text states how to read the simplified map; Figure~\ref{fig:evaluation-benchmark-map-full} lists every coded method--benchmark link for spot-checking coverage and benchmark labels.

The maps draw on the evaluation methods analyzed in Section~\ref{sec:evaluation}: 28 methods, 100 method--benchmark links, and 68 normalized benchmark nodes in the complete appendix figure.
Each link records that a method reports results on, or is explicitly evaluated on, a named benchmark in its source paper.
A method may contribute several links when it reports several benchmarks, but at most one link per method--benchmark pair after light name normalization.
During manual review, paper-specific source notes guide whether a benchmark name is retained and which evaluation family it supports.

The main-text figure drops links whose benchmark appears for only one method, leaving 47 links, 15 shared benchmark nodes, and 21 methods with at least one shared benchmark.
It therefore emphasizes comparable empirical surfaces rather than the full coverage set.
Figure~\ref{fig:evaluation-benchmark-map-full} restores singleton benchmarks and is intended as a completeness check, not a separate taxonomy.

We code benchmark nodes at the benchmark-series level.
Name normalization groups related releases under a shared series label.
We exclude links supported only by local reruns or other evidence outside a method's originally reported benchmark set.

\paragraph{Benchmark artifact types}
Each benchmark node is assigned one artifact type for visual encoding in both figures.
The three classes follow the legend in Figure~\ref{fig:evaluation-benchmark-map}.
\emph{Fixed response or judgment sets} package candidate responses, pairs, rankings, or step labels for direct judge comparison.
\emph{Task benchmarks requiring generated candidates} supply tasks, references, or checks whose reported results require generating candidates before judging.
\emph{NLG evaluation datasets} provide open-text outputs with human quality, factuality, or facet labels for evaluator meta-evaluation.
Candidate extensions in the plotting data are rendered with muted styling when they appear in the simplified main-text figure.
For boundary cases excluded from the NLG-evaluation axis, see Section~\ref{sec:evaluation-nlg}.

\end{document}